\RequirePackage{fix-cm}
\RequirePackage{amsmath}
\documentclass[twocolumn]{svjour3}          
\smartqed  

\usepackage{amsmath,amssymb,amsfonts}
\usepackage{algorithm,algorithmic}
\usepackage{graphicx}
\usepackage{textcomp}
\usepackage{tikz}
\usepackage{booktabs}
\usepackage{multirow}
\usepackage{wrapfig}
\usepackage{float}
\usepackage[numbers]{natbib}

\usetikzlibrary{shapes.geometric, arrows}
\tikzstyle{startstop} = [rectangle, rounded corners, minimum width=in, minimum height=1cm,text centered, draw=black, fill=red!20, rounded corners]
\tikzstyle{io} = [trapezium, trapezium left angle=70, trapezium right angle=110, minimum width=1in, minimum height=1cm, text centered, draw=black, fill=blue!20]
\tikzstyle{process} = [rectangle, minimum width=1in, minimum height=0.25in, text centered, draw=black, fill=white!20, rounded corners]
\tikzstyle{decision} = [diamond, minimum width=1in, minimum height=0.3in, text centered, draw=black, fill=green!20, rounded corners]
\tikzstyle{arrow} = [thick,->,>=stealth]

\def\BibTeX{{\rm B\kern-.05em{\sc i\kern-.025em b}\kern-.08em
    T\kern-.1667em\lower.7ex\hbox{E}\kern-.125emX}}
    
\begin{document}
\title{
Protein Crystallization Segmentation and Classification Using Subordinate Color Channel in Fluorescence Microscopy Images    
}

\titlerunning{Protein Crystallization Segmentation and Classification Using Subordinate Color Channel}  

\author{Truong X. Tran \and Marc L. Pusey \and Ramazan S. Aygun}

\institute{Truong X. Tran and Ramazan S. Aygun \at
              Data Media Lab, Computer Science Department, The University of Alabama in Huntsville, USA.
              \email{\{txt0002,aygunr\}@uah.edu}           
           \and
           Marc L. Pusey \at
              XpressGenes, Inc, Huntsville, Alabama, USA. \\
              Chemistry Departement, The University of Alabama in Huntsville, USA.  
              \email{mlp0041@uah.edu}
}

\date{}

\maketitle

\begin{abstract}
The accuracy of detecting protein crystals for fluorescence microscopy images is very critical for high throughput and automated systems. Although the trace fluorescent labeling method could highlight protein crystals, reflection and emission from the fluorescence dye is not always due to crystal regions. Therefore, the analysis of the peak wavelength in the emission spectra of a fluorophore may not always yield effective results. In this paper, we show that using the subordinate color intensity corresponding to longer wavelengths than the peak wavelength of the emission spectra could improve the accuracy of protein crystal detection. 
Hence, we have built a segmentation method based on the percentile intensity of the subordinate color for trace fluorescently labeled (TFL'd) protein crystallization trial images. Compared to using the dominant color channel, our segmentation method on subordinate color channel was able to reduce the misclassification rate of likely-leads or crystals as non-crystals by the percentage of from $9.71\%$ to $2.02\%$ depending on the classifier. Similarly, the accuracy of classifiers were increased by the percentage of from $1.77\%$ to $5.53\%$. Our method reached around $94\%$ accuracy while keeping misclassification of likely-leads and crystals as non-crystals below $1\%$. Moreover, to evaluate the generalizability of our method, we have conducted new wet lab experiments on two proteins, Concanavalin A (Con A) and Ab inorganic pyrophosphate  (AbIPPase), and the misclassification rate was below $1\%$. Our experiments show that using the subordinate channel may be more helpful for TFL'd protein trial image classification.

\keywords{Fluorescence Image Analysis \and Protein Crystallization\and Image Segmentation \and Classification}

\end{abstract}

\section{Introduction}
\begin{figure}
\centerline{\includegraphics[width=0.5\textwidth]{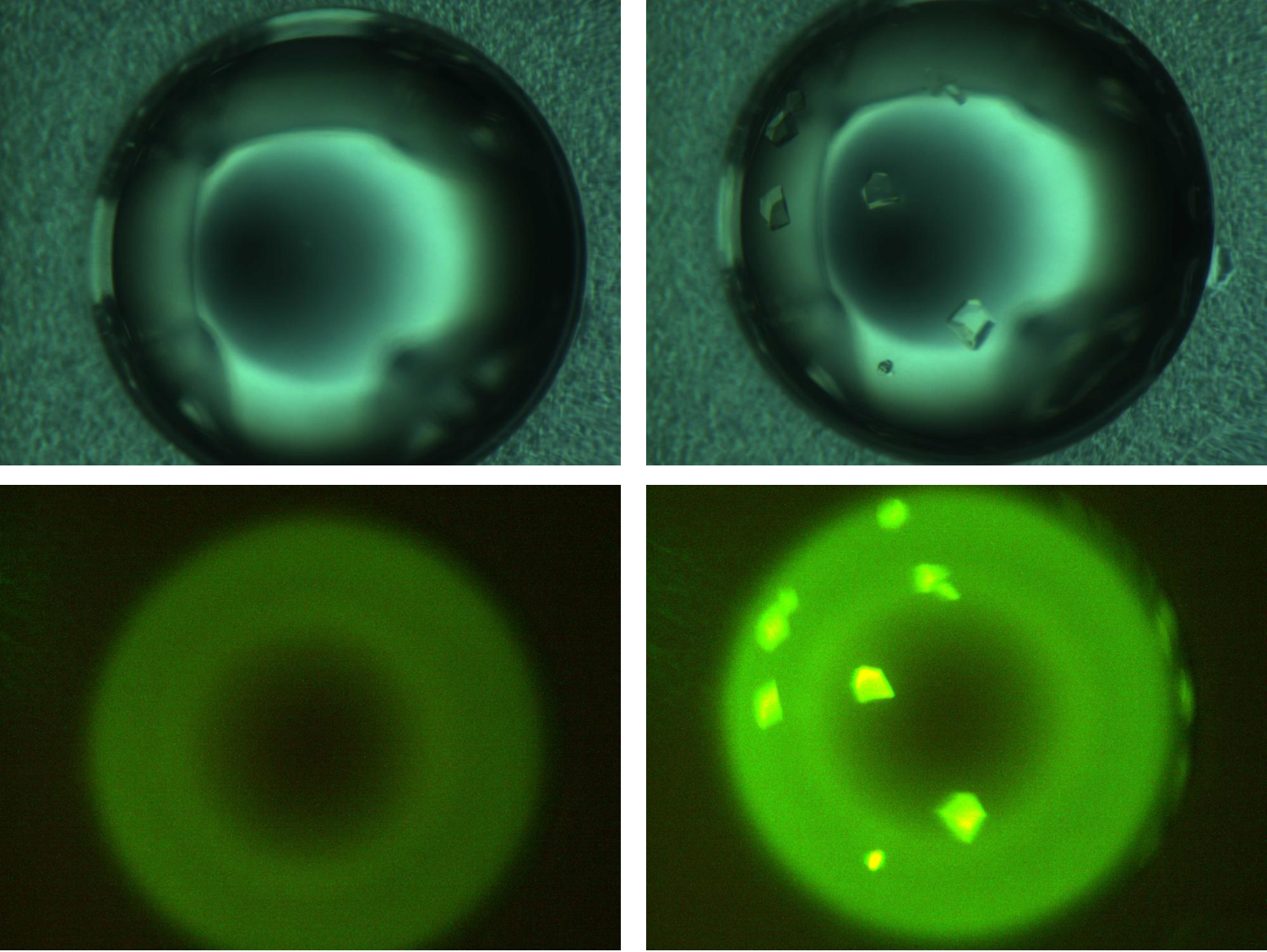}}
\caption{White light (top) and fluorescent images of non-crystal (left) and crystal (right) sample wells.}
\vspace{-10pt}
\label{fig:TFL}
\end{figure}

High-throughput systems could handle both setting up experiments and evaluating outcomes to aid crystallographers in conducting a large number of experiments. Image acquisition, analysis, and classification are major components of evaluating thousands of trial images on a daily basis. Trace fluorescent labeling (TFL) utilizes fluorescence dye to identify crystals that would have high intensity when exposed to light at the corresponding wavelength. This technique allows to acquire a clear view of wells in plates thus enabling to build a standalone imaging and classification system for analyzing plates \cite{ForsythePusey2006:TFL,Pusey2015:TFL,Sigdel2013:Real-Time,sigdel2015crystpro}. Fig. \ref{fig:TFL} shows images of sample wells that were taken under white light and fluorescence illumination. 

Image analysis techniques typically focus on the emission spectra of the fluorescence dye to identify crystals and the phases of crystallization. Nevertheless, reagents in cocktail solutions may also reflect light. Fluorescence dye molecules may float in the solution until they attach to a protein crystal, and these may complicate the imagery analysis of crystallization process. Traditional techniques naturally focus on wavelengths around the excitation/emission wavelengths of the fluorescence dye and have to deal with reflections and emissions from non-crystal areas in the solution. Although reflection or emission is possible from non-crystal areas in solutions at dominant wavelengths, emission or reflection is likely to be lower at farther wavelengths from the dominant wavelength. As the fluorescence dyes attach to protein crystals, their emission at farther wavelengths will also be significant due to their density and this could help to separate crystal regions from non-crystal regions.

\begin{figure}[b]
\centerline{\includegraphics[width=0.45\textwidth]{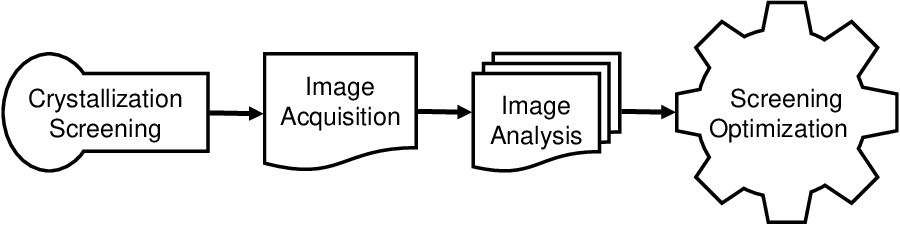}}
\caption{Basic analysis pipeline of protein crystallization (refer to \cite{Tran2017:subordinate}).}
\vspace{-15pt}
\label{fig:pipeline}
\end{figure}

In this paper, we  analyze whether processing based on subordinate channel corresponding to farther wavelengths than the peak wavelength of the emission spectra could provide better results or not than methods relying on the dominant color channel (based on the peak wave length). In our previous work, we have utilized subordinate channel while building and evaluating a new classification technique, the else-tree classifier \cite{tran2018else}. In \cite{Tran2017:subordinate}, we have built a framework for determining the best parameter set while classifying protein crystal images. Using this framework, in this paper we explain our segmentation method based on subordinate channel that could improve the accuracy of classifiers while reducing the misclassification rates compared to using the dominant color channel.
We analyze percentile based thresholding methods to compare the performance of using features based on the subordinate color channel and dominant color channel. After the features are extracted, we use a number of classifiers to build a learning model and show how those features obtained from the corresponding channels affect the accuracy.
Compared to our work in \cite{Tran2017:subordinate}, this paper adds more details on the analysis and experimental results of using dominant and subordinate color channels. 
Moreover, to show the generalizability of our method, new wet-lab experiments were conducted for Concavanlin A (Con A) and Ab inorganic pyrophosphate (AbIPPase) proteins, and we show that it is possible  to reduce the misclassification rate of crystals as non-crystals below 1\% using the subordinate color channel.       

This paper is organized as follows. The following section explains the related work on classification of protein crystallization trial images. Section III provides background on TFL, explains the data set used building our learning model, reviews the relevant thresholding techniques used in previous research, introduces the region features extraction of the trial images, and presents our classification model building framework. 
Section IV covers our method including the concept of subordinate color, the analysis of TFL image intensity, and the subordinate percentile intensity thresholding method. Experiments and results are discussed in Section V. Discussion of experimental results is provided in Section VI. The last section concludes our paper.

\section{Related Work}\label{Related}
As shown in Fig. \ref{fig:pipeline}, there are four major steps in the analysis of protein crystallization experiments include crystallization screening, image acquisition, image analysis, and screening optimization. In the screening stage, the protein solution is prepared and dropped into a well-plate. The plates are scanned in the image acquisition stage and then the captured images are stored in a database. During the image analysis stage, the images are classified or given a numerical score indicating the phase of crystallization. The relevant screens can be selected and optimized for getting higher quality crystals.   

There has been significant work on image segmentation and classification to increase the quality of protein crystallization analysis.  Since high-throughput screening systems are automated systems, it is critical to reduce the possibility of missing crystallization outcomes. In 2002, Spraggon et al. classified the result of protein crystallization experiments based on analyzing texture features and edge detection of trial images \cite{Spraggon2002:Computational}. They used a self-organizing neural network for the classification model. The model accuracies were from 47\% to 82\% based on different categories (from clear drop to mountable crystal). In 2004, Zhu et al. analyzed the trial images using geometric and texture features for classifying crystal and non-crystal screening results using a support vector machine and automatic decision tree model. They reported the lowest false-negative rate of 9.6\% while the best false positive rate is 14.6\% \cite{Zhu2004:Classification}. Bern et al. (2004) analyzed the gradient and geometric features of the in-focus region in the image of the sample solution \cite{Bern2004:AutomaticClassification} and classified images into five classes (‘empty’, ‘clear’, ‘precipitate’, ‘microcrystal hit’ and ‘crystal’) using the decision tree algorithm. The testing results yielded about 12\% false negatives and 14\% false positives. In 2006, Pan et al. extracted the texture and Gabor wavelet decomposition from the trial images \cite{Pan2006:AutomatedclassificationusingSVM}. Pan et al. also used the support vector machine to classify three categories of clear drop, precipitate, and crystal results. They obtained a low false negative rate of under 4\%. However, the false-positive rate was about 40\%.            

In addition, there were many studies on analyzing a larger set of features or using the image itself as the input for training classification models. In 2008, Po et al. detected the regions of interest and applied multi-scale Laplacian pyramid filters and histogram analysis on the trial images \cite{PO2008:Leveraginggenetic}. Po et al. combined the genetic algorithms with neural network classification models. Their method produced 99\% true negative and 88\% true positive rate. In 2010, Cumbaa et al. obtained the set of basic image statistics, energy, Euler number, Radon-Laplacian, Sobel-edge, microcrystal, and GLCM features \cite{Cumbaa2010:ProteinCrystallizationGrid}. They trained a random forest classification model to classify three categories of crystal-bearing images,  precipitate images, and clear drops. The classification accuracy per category is 80\%, 89\%, and 98\% respectively. As an application of deep learning, Bruno et al. (2018) \cite{bruno2018classification} presented a machine learning model that could correctly recognize 94\% of test images while categorizing images into four categories as crystals, precipitate, clear, and other outcomes. Their   deep neural network was trained on half a million crystallization trial images.

In our works, we have analyzed and classified TFL'd protein crystallization trial images for our automated fluorescence microscopy system. Sigdel et al. (2013) \cite{Sigdel2013:Real-Time} obtained the intensity and region features to classify three classes (non-crystal, likely lead and crystal) of the trial images. We got 88\% overall accuracy with only 1.2\% misclassifying crystal as non-crystal rate when using using multiple perceptron neural network (MLP) and max-class ensemble models. In 2014, Sigdel et al. \cite{Sigdel2014:GeometricFeaatures} used decision tree model to classify four crystal image categories (needles, small crystals, large crystals, and other crystals) based on geometric features. In 2017, Sigdel et al. (2017) \cite{Sigdel2017:Featureanalysis} evaluated several combinations of features sets with different classification models. We also applied feature reduction and selection techniques to improve classification performance. In our analysis, we got 96\% accuracy with 87\% sensitivity on three classes of non-crystals, likely leads and crystals using a large feature set consists of intensity features, region features of Otsu's thresholding, green percentile thresholding G90 and G99, histogram features, and graph features.     

Most of the research studies on categorizing protein crystallization trial images captured under white light. These studies used different feature extraction techniques and classification models on varying number of protein crystallization categories. Although it is difficult to compare the performance of these methods, in general, the researchers claim an accuracy around 90\% with misclassification rate of crystals around 10\%.
We should note that  the research studies usually report their results based on the light source the images are captured.  
Previously, we have built the else-tree classifier \cite{tran2018else} and a framework that determines the best features, methods, parameters, and classifiers for fluorescence images \cite{Sigdel2017:Featureanalysis,Tran2017:subordinate}. Recently, Klijn and Hubbuch (2019) \cite{klijn2019time} present their work on combining the end points and time-dependent features using 3 different light sources for classifying the protein phases. The combination of different light sources have a better balance accuracy result than a simple complex classification models. However, in the literature, for fluorescence microscopy, we have not seen the comparison of classification based on dominant color wavelength and subordinate color channel.  Hence, it is important to analyze how subordinate color channel processing impacts classification accuracy.

\section{Trace Fluorescent Labeling and Image Classification}\label{Background}
Protein crystallization experiments utilize characteristics of a specific protein and chemical solution to produce successful screening results. Robotic high-throughput systems help to carry out many experiments with fast trial image capturing capability. The analysis of the trial images is essential for a better understanding and observing the screening result. Altogether, the goal is to increase the success of the crystal growth process. This section reviews the basics of our protein crystallization process, our TFL2756 trial image data set, our previous image segmentation method, the summary of our region features extraction, and the general framework to build and evaluate our classification model.    

\subsection{Trace Fluorescent Labeling Protein Crystallization Analysis}\label{TFL procedure}
In our wet lab, we applied the Trace Fluorescent Labeling (TFL) method to rapidly reveal the appearance of the crystal in a trial solution. Thus the crystals of labeled protein have a high concentration of fluorescence dye and they can be visualized by fluorescence imaging  \cite{ForsythePusey2006:TFL,Pusey2015:TFL}. 

\begin{figure}
\centerline{\includegraphics[width=0.25\textwidth]{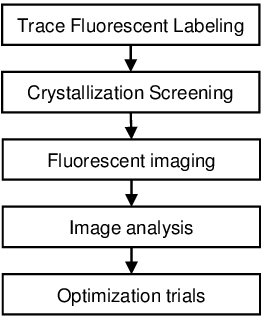}}
\caption{Stages of protein crystallization analysis using TFL (refer to \cite{Tran2017:subordinate}, \textcopyright2017 IEEE.).}
\vspace{-15pt}
\label{fig:TFLAnalysis}
\end{figure}

Fig. \ref{fig:TFLAnalysis} shows the major steps in our protein crystallization experiments. We first label the protein with fluorescence dye. Then, we prepare the sample solution during the screening phase. After the preparation, the sample plates are carefully sealed and stored in an incubator. The plates are scanned periodically. Normally the first scan starts right after setting up the sample plates and follows by either a daily or weekly period. For capturing the image of each trial well of the  plates, we place each of them in the dark chamber of a fluorescence microscope. Then, the microscope automatically scans all wells of the plate. In the image analysis phase, each trial image is visualized, segmented, and classified with scores. The score is an evaluation outcome of the experiment with respect to the crystallization phase ranging from a clear drop non-crystal to a big 3D crystal. There are many other phases of crystallization as well. All of the image analysis processes are computer-supported. The experimenter is also allowed to review the trial plates to verify the computer analyzed results. The analysis results are used for the later optimization experiments to optimize the leading crystalline conditions.

\begin{figure}[t]
\centerline{\includegraphics[width=0.4\textwidth]{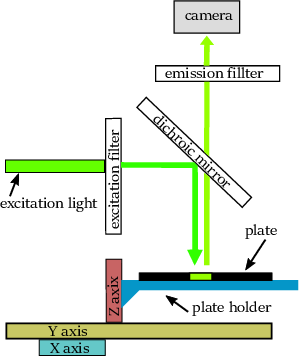}}
\caption{Layout of fluorescence microscope system.}
\vspace{-10pt}
\label{fig:Microscope}
\end{figure}

We use our custom Crystal X2 fluorescence microscope \cite{Sigdel2013:Real-Time} as the main instrument for imaging and analysis the crystallization sample plates. The layout of Crystal X2 is presented in Fig. \ref{fig:Microscope}. The microscope is composed  of an illumination component, optical component, automated control mechanism, and computer-aided software. The microscope uses LED illuminators to minimize the changing of temperature when capturing the sample images. The optical component contains a low pass excitation filter (Omega XF1080), a dichroic mirror (Omega XF2012), a high pass emission filter (Omega XF3021), and an imaging camera. The control mechanism automatically slides the plate in the X and Y direction to locate the  wells based on the plate configuration\cite{Pusey2015:TFL}. When a well is located, the software controls the camera to capture the trial image. 

During image capturing, the light (LED) goes through the excitation filter and is reflected by the dichroic mirror to focus on the sample well. The emitted fluorescence light from the well sample solution is collected by the same dichroic mirror, then gone through the emission filter toward the camera lens. The software controls the camera to capture the trial image from the well. Finally, all trial images are stored in the database together with other relevant experiment information \cite{Pusey2015:TFL,Sigdel2013:Real-Time}.

\subsection{TFL2756 data set}
\begin{figure}[t]
\centerline{\includegraphics[width=0.5\textwidth, trim={0 0 0 0.261in},clip]{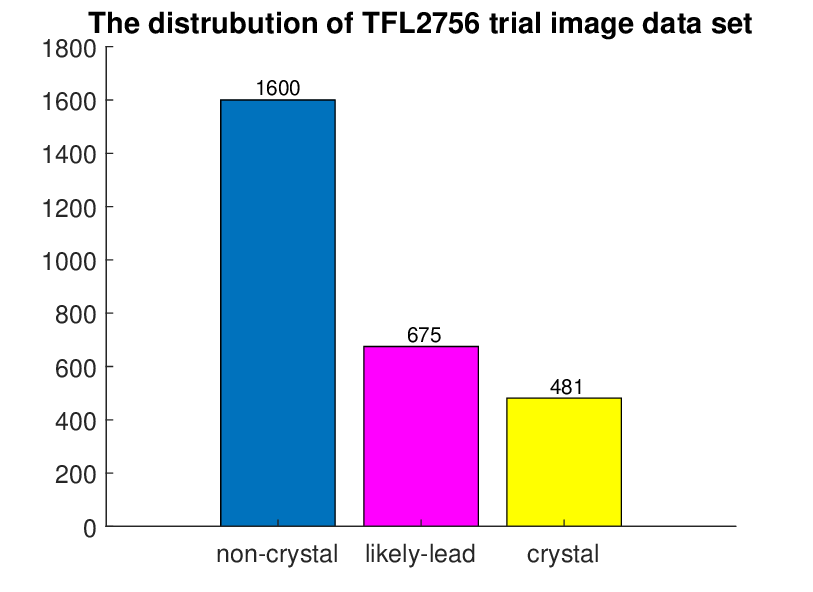}}
\caption{Distribution of TFL2756 data set.}
\vspace{-10pt}
\label{fig:TFL2756}
\end{figure}

\begin{figure}[t]
\centering

\includegraphics[width=\linewidth]{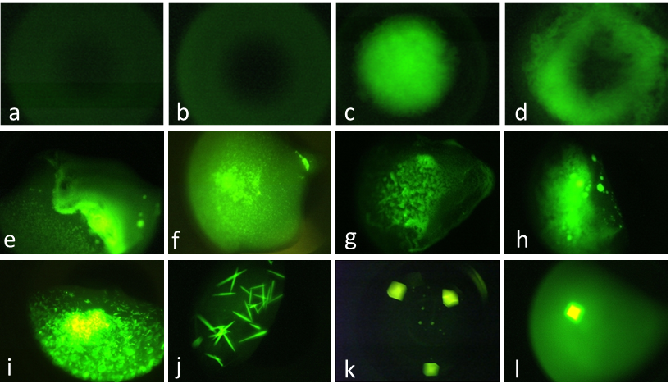}
\caption{Non-crystal (a-d), likely-lead (e-h), and crystal (i-l) sample images.}
\label{fig:SampleCategories}
\end{figure}

In our research, we gathered a data set of 2756 trial images from different TFL experiments. The crystallographer expert manually reviewed and classified each image into three categories based on Hampton's scale  \cite{Sigdel2017:Featureanalysis}. The categories and frequencies of images in this study are displayed in Fig. \ref{fig:TFL2756}. The bar chart shows the number of images in non-crystal, likely-lead, and crystal categories. Around 60\% of the images are non-crystal. The images which do not contain clear crystals but whose conditions may lead to crystals are labeled as likely-leads.  
The use of an intermediate class (likely-lead) may reduce the costly mistake of missing crystal without requiring to review all non-crystals \cite{Sigdel2013:Real-Time}. Fig.~\ref{fig:SampleCategories} shows sample images of the TFL2756 data. Each image has dimensions of $320\times240$, 32-bit depth, and 96 dpi resolution. 

\subsection{TFL'd Image Segmentation based on Dominant Color Wavelength}
The intensity-based analysis is suitable for TFL images since protein crystals have a higher intensity than the background. Our TFL imaging uses green LED light (530nm) as the excitation source, thus the dominant color channel in the trial images is green. Therefore, in our previous work, we used a green percentile intensity thresholding method to segment the images \cite{Sigdel2013:Real-Time}. We have used green percentile thresholding while developing our decision-tree based binarization method \cite{Dinc2014:DTbinarize} and supervised thresholding method to choose the best binarization method \cite{Dinc2017:Super-Thresholding}.
Moreover, we have conducted extensive studies using this threshold for generating geometric features \cite{Sigdel2014:GeometricFeaatures} and comprehensive evaluation of feature sets \cite{Sigdel2017:Featureanalysis}.
In our studies, G90, G95, G97, G99, and G100 are five thresholding that generated proper results. For instance, in the case of G90 thresholding, a threshold value T\textsubscript{G90} is computed as the 90\textsuperscript{th} percentile intensity of the green channel. In this case, 90\% of the pixels in the image have green intensity less than the threshold T\textsubscript{G90}. The green percentile intensity thresholding also uses a grayscale (t\textsubscript{min}) threshold to require the pixels of interest to have some minimal intensity if crystallization occurs. 

For image binarization, a pixel is considered as foreground if its overall gray intensity is greater than t\textsubscript{min} and green intensity greater than T\textsubscript{G90}. On the other hand, the maximum green intensity thresholding, G100, uses T\textsubscript{G100} - the maximum green intensity among all pixels in the image.  In this case, the foreground constitutes all pixels with the green component intensity equal to T\textsubscript{G100} and gray level larger than t\textsubscript{min}. In the rest of the paper, to indicate that the green channel is the dominant color channel, we will add prefix {\em Dom} in front of green percentile thresholding. For example, $DomG90$ is the same as $G90$ but it indicates that green is the dominant color channel. 

\subsection{Region Features Extraction}\label{section:feature-extraction}
There is a trend to increase the number of image features to improve the classification performance. However, the use of irrelevant features may reduce the performance of some classifiers. The comparison and analysis of features for protein crystallization were present in one of our work \cite{Sigdel2017:Featureanalysis}. 
In this paper, we use  connected region features from the original trial image and general binary image features from its binarized image. For the binary image features, we focus on the intensity of the image since the crystals have the highest emission compared to other precipitates. For the region features, we focus on the shape and size of the large blobs. 

For each binary image, 7 features are extracted: the ratio of foreground to image size, the ratio of convex hull area to image size, foreground average intensity, standard deviation (stdev) of foreground  intensity, background average intensity, standard deviation for background  intensity, and the number of connected regions (blobs). 

We used connected labeling to segment the blobs in the binary image. When the regions were recognized, we selected the top $5$ largest blobs in each image and extract the region features in the original image. There were 9 features extracted for each region: average region intensity, stdev of region intensity, region area, region filled area, region perimeter, equivalent circular diameter, convex hull area, eccentricity, and region’s extent \cite{Sigdel2013:Real-Time,Sigdel2017:Featureanalysis}. As a result, 52 features were extracted from each image:
\begin{equation}
    features = 
    \begin{array}{l}
    \textit{7 binary features}
    \\
    + (\textit{9 region features}) \times \textit{(5 large regions)}
   \end{array}
\end{equation}

\subsection{Building Classification Model Using Image Segmentation Features}

\begin{figure}[!t]
\centerline{\includegraphics[width=0.3\textwidth]{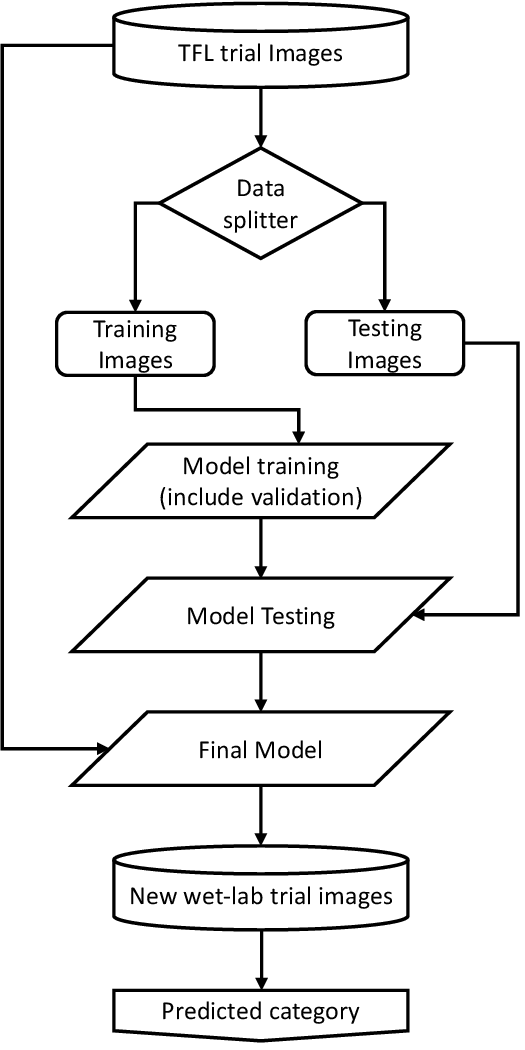}}
\caption{The framework of classifying TFL'd protein crystallization images.}
\label{fig:Model-Building}
\end{figure}

The framework for building our classification model is provided in Fig.~\ref{fig:Model-Building}. There are three major stages in the framework: i) model training and validation, ii) model testing, and ii) final model evaluation. These steps are summarized as follows:

\begin{itemize}
\item First, the TFL2756 data is randomly divided into training and testing data sets by stratified sampling. 

\item In the model training and validation stage, the training images are segmented by a binarization method. Then, the region features are extracted. The classification model is trained on the extracted features set. The model also is evaluated in the training process, such as using 10-fold cross-validation.  

\item In the model testing stage, the testing data set is used to evaluate the classification model. The extracted features in each testing image are fed to the classification model and the prediction result is compared with the actual class label.

\item Finally, in our study, we use whole TFL2756 data to train our final classification model. The final model is used to classify the new wet-lab protein crystallization images. 
\end{itemize}

\section{Proposed Method: Protein Crystallization Segmentation and Classification Using Subordinate Color Channel}
\label{Proposed}
We develop a new method to analyze the protein crystallization trial images using a subordinate color channel rather than the dominant color in the fluorescence microscopy images. We first introduce the concept of subordinate color and the basics of TFL image intensity analysis. Then we explain our subordinate percentile intensity thresholding method for segmenting TFL images in the subordinate color channel. 

\subsection{Subordinate Color Channel in Fluorescence Emission}
In fluorescence microscopy, the sample is usually labeled with a fluorophore that could emit light when illuminated.  
In order to achieve an image with maximum fluorescence intensity, the emission wavelengths near or at the peak of the emission are selected for detection. The selected wavelengths create the dominant color in the image achieved by the fluorescence microscopy. In most cases the higher the intensity, the better the image analysis result. However, the analysis of TFL'd protein crystallization trial images in the dominant color channel is not always easy. 
Although protein crystals generally have the highest intensity in an image, sometimes there might be bright spots that may not be crystals \cite{Pusey2015:TFL}. Moreover, in some images, crystals appear within a large lower intensity area. In some images, crystals might be out of focus and blurry. Since segmentation and classification of those images are challenging, this problem motivated us to analyze other wavelengths than the the emission wavelength.

\begin{figure*}[t]
\centering
\includegraphics[width=0.7\linewidth]{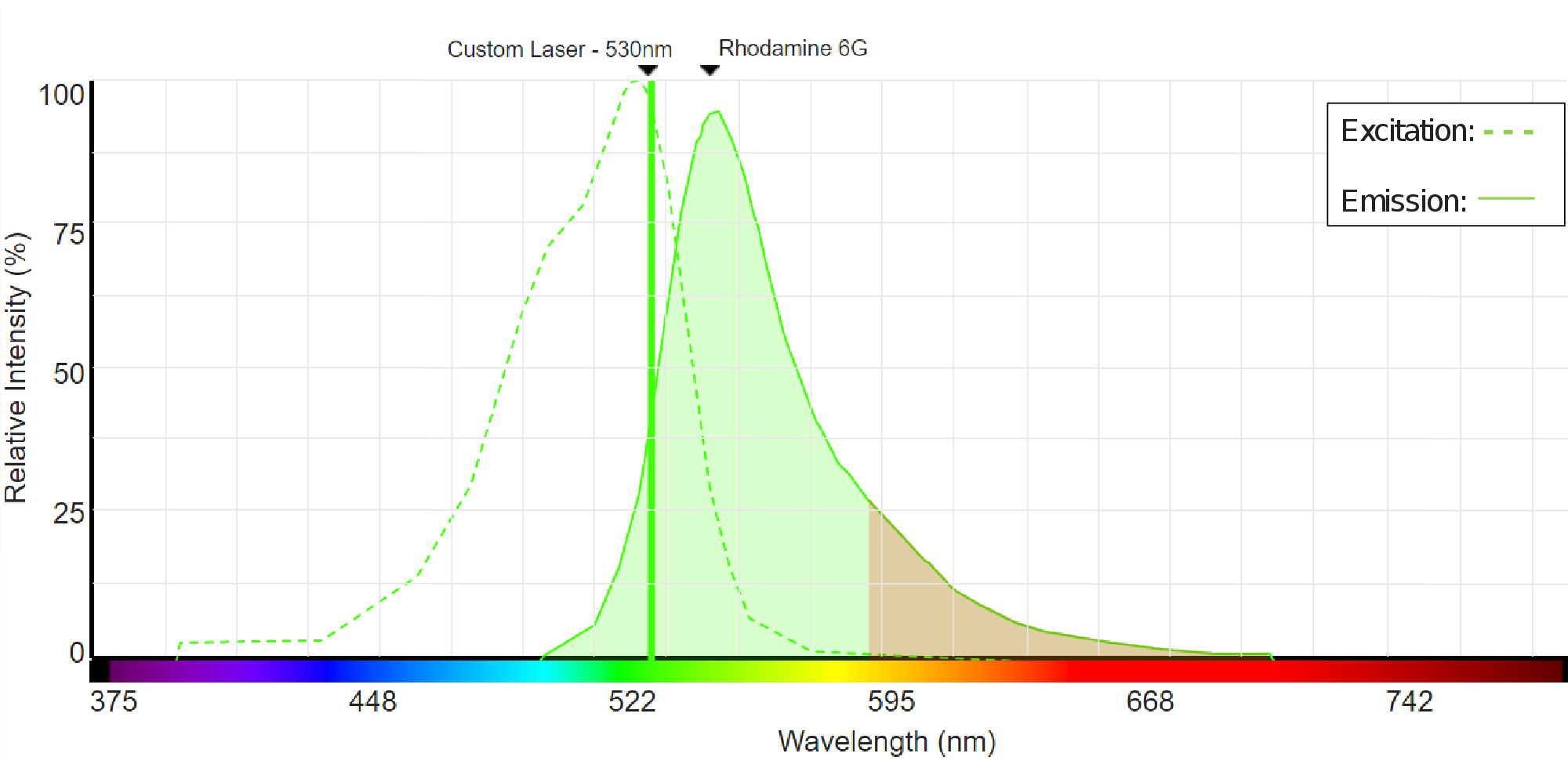}
\caption{Fluorescence spectra (simulated using SpectraViewer tool \cite{SpectraViewer}) when illuminating Carboxyrhodamine 6G - Succinimidyl Ester (CR-6G) with a 530nm Laser light. }
\label{fig:Spectra}
\end{figure*}

Consider the spectrum of excitation and emission lights for Carboxyrhodamine 6G - Succinimidyl Ester (CR-6G, C6157)  illustrated in Fig.~\ref{fig:Spectra}. The CR-6G has excitation and emission spectra in between those of fluorescein and tetramethylrhodamine and have higher fluorescence quantum yield (the ratio of absorption to emission photons) than tetramethylrhodamine conjugates. With the peak absorption of CR-6G at 525nm, we use a 530nm laser light in our microscope as the illumination source. As shown in the figure, the peak intensity of the emission when illuminated by a 530nm light is around 550nm wavelength. However, CR-6G also emits light for wavelengths greater than 590nm despite having lower intensity in images. If the far right side of Fig. \ref{fig:Spectra} for wavelengths from 595-700nm is considered as a long tail, this side has intensity values in the red channel of RGB images. The relative intensity interval is from $\sim$28\% to 2\%.  

We consider the emitted light corresponding to the longer wavelength (red channel in our case) with low intensity as the subordinate color channel in our fluorescence image. In Fig. \ref{fig:Spectra}, it can be seen that there is not much emission for wavelengths less than $500nm$ and do we do not expect any intensity for these wavelengths corresponding to bluish colors.

\begin{figure*}[t]
\centering
\includegraphics[width=0.7\linewidth]{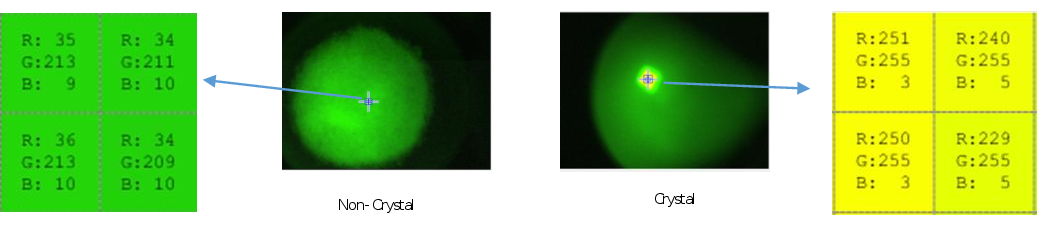}
\caption{Inspection pixel values for non-crystal and crystal regions. }
\label{fig:IntensityPeeking}
\end{figure*}

\begin{figure}[t]
\includegraphics[width=\linewidth]{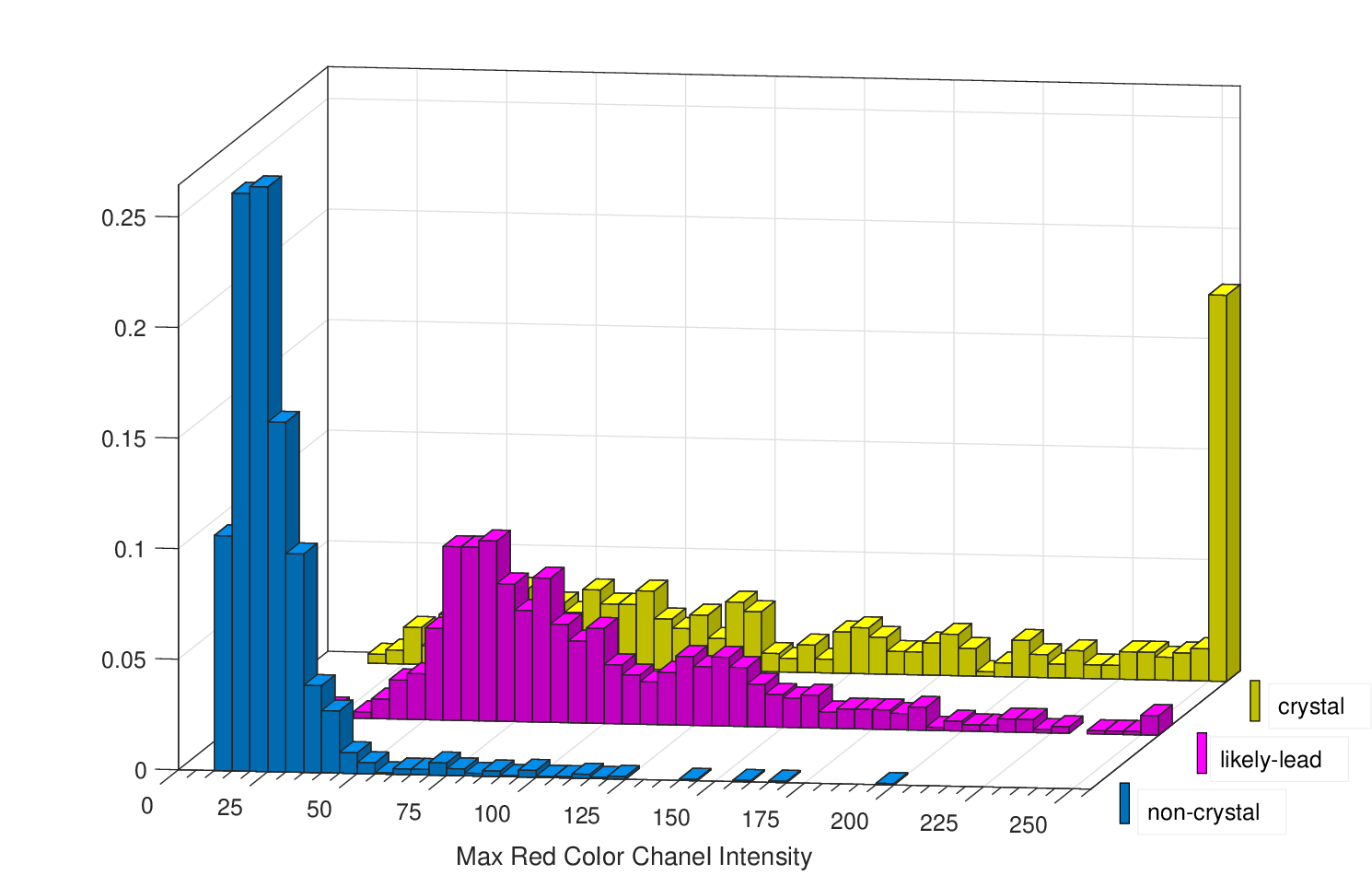}
\caption{Histogram of maximum red intensity for all three image categories.}
\label{fig:HistMaxRed}
\end{figure}
\begin{figure}[t]
\includegraphics[width=\linewidth]{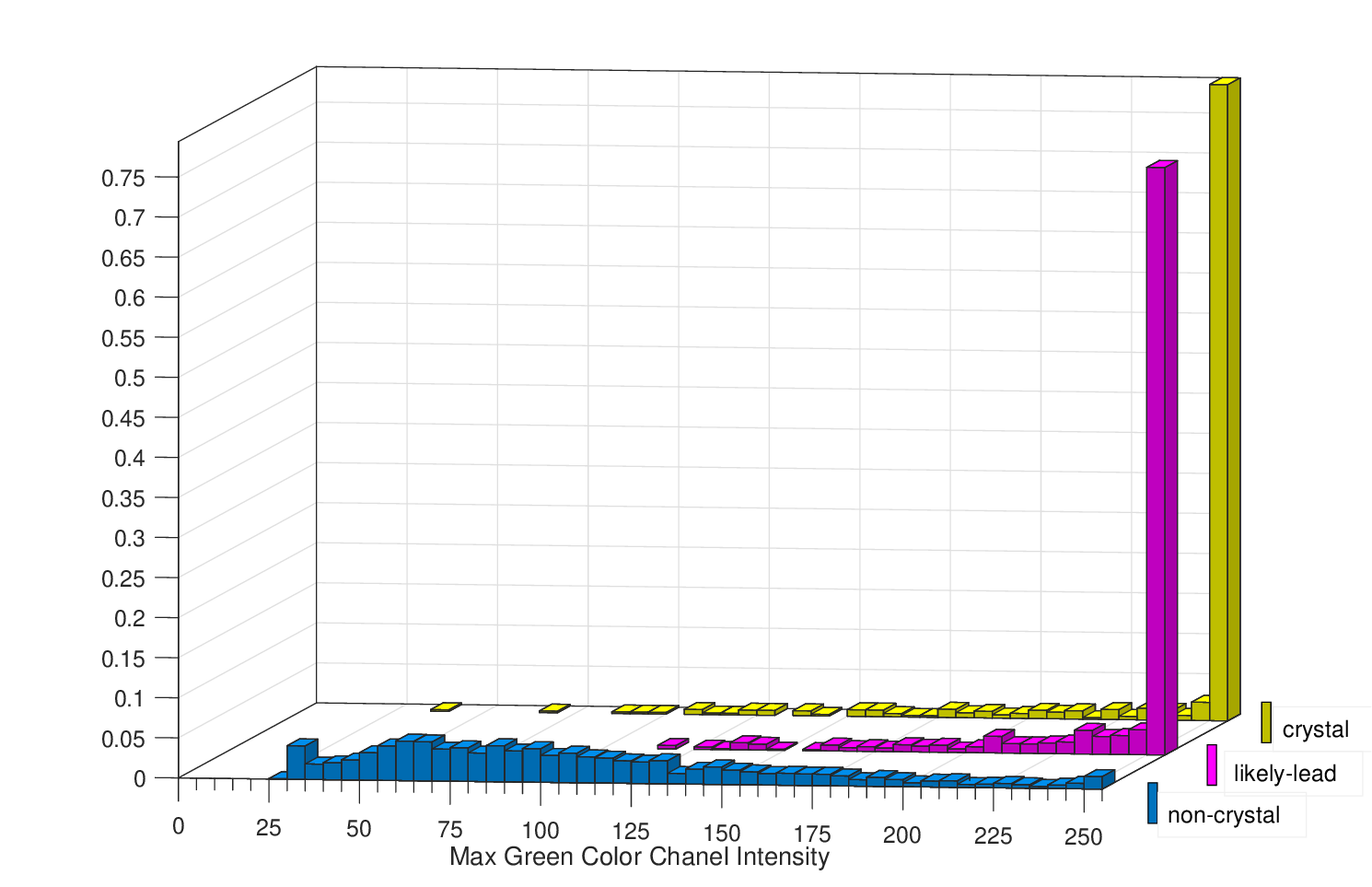}
\caption{Histogram of maximum green intensity for all three image categories.}
\label{fig:HistMaxGreen}
\end{figure}

\begin{figure}[t]
\includegraphics[width=\linewidth]{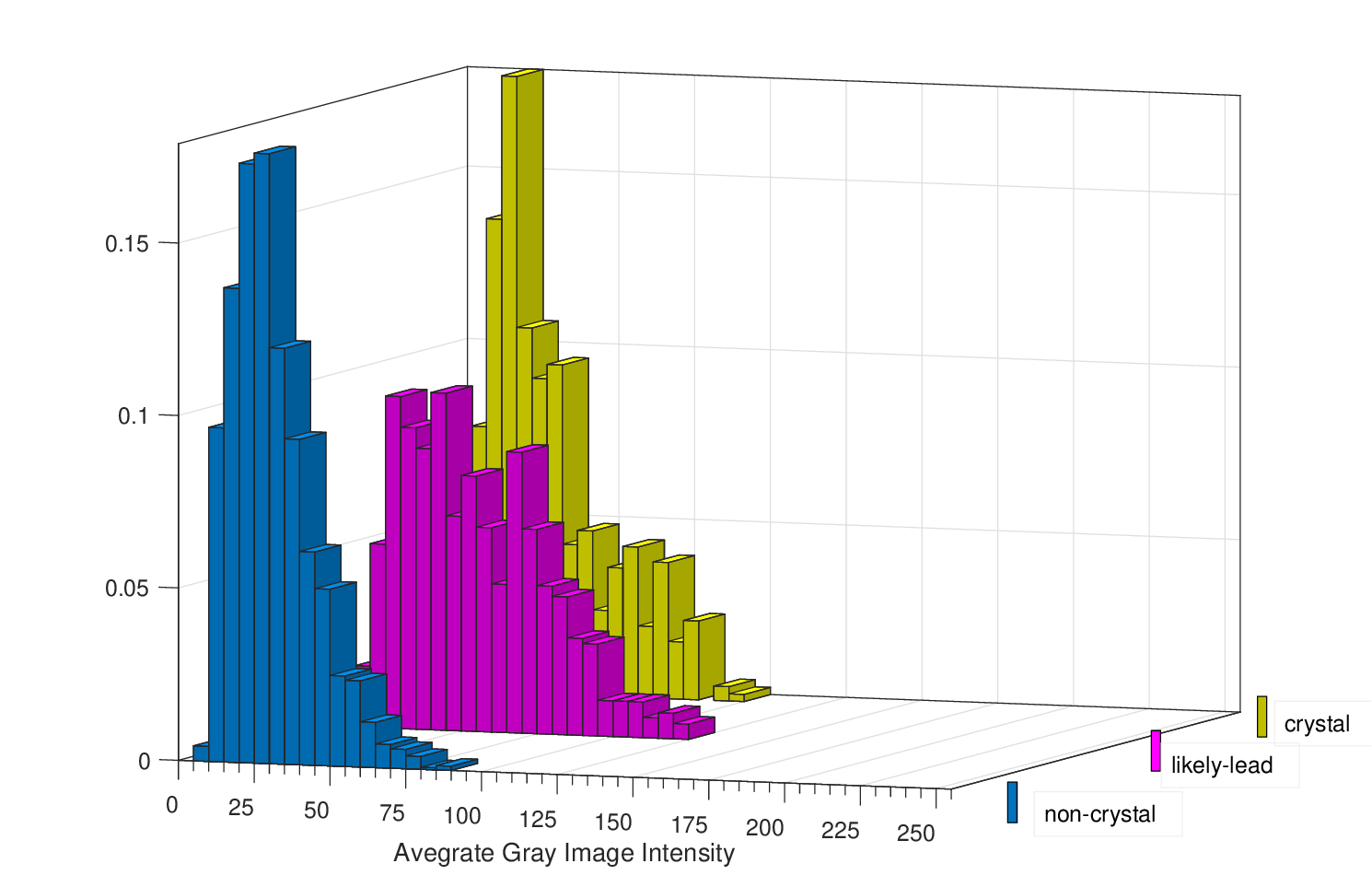}
\caption{Histogram of average gray image intensity for all three image categories.}
\label{fig:HistAvgGray}
\end{figure}
\begin{figure}[t]
\includegraphics[width=\linewidth]{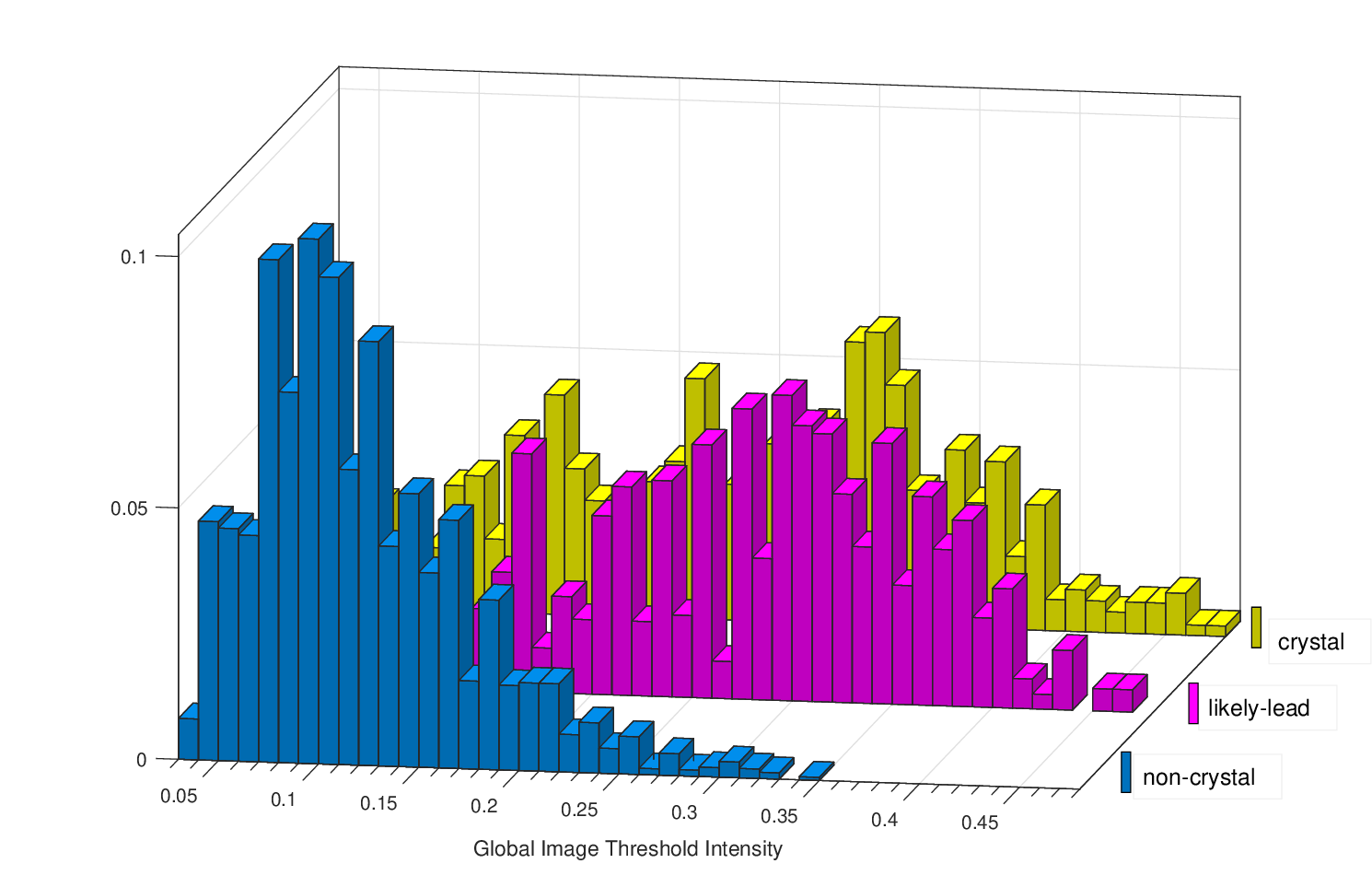}
\caption{The global intensity histogram threshold using Otsu's method for all three image categories.}
\label{fig:HistOGray}
\end{figure}

\subsection{Color Intensity Analysis of The Protein Crystallization Trial Images}
We propose using other color channels based on the filter used on the microscope rather than using the primary color to analyze the protein crystallization trial images. We hypothesize that the analysis of other color channels may provide better results than the analysis of the primary color channel.

In the TFL2756 data sets, the green color is dominant and the red color is subordinate. Fig. \ref{fig:IntensityPeeking} displays the pixel values at crystal and non-crystal regions in RGB trial images of a clear drop sample (unsuccessful trial) and a protein crystal (successful trial). In this example, green color has a high intensity value ($>210$) in both crystal and non-crystal images. If non-crystal and crystal images are analyzed, the green intensity variation is not very significant for crystal images and non-crystal images since in another image it is possible to have a crystal region with a lower green intensity than in a non-crystal region in another image. It is desirable that the intensity change is significant between a crystal region and non-crystal region across images. If the subordinate red color is considered for comparison, the change in red color intensity (around $34$ in the non-crystal image whereas $>225$ in the crystal image) is highly different and it could serve as a better candidate to distinguish crystal and non-crystal regions.

To further analyze the subordinate color channel, for each image ($I_{h\times w}$) in non-crystal, likely-lead, and crystal categories, the maximum green and red intensity values are taken as follows:
\begin{equation}
I_{max}^{Green} = MAX_{1\leq i \leq h,1\leq j \leq w}(I_{Green}(i,j)),
\label{eq-max-intensityGreen}
\end{equation}

\begin{equation}
I_{max}^{Red} = MAX_{1\leq i \leq h,1\leq j \leq w}(I_{Red}(i,j)).
\label{eq-max-intensityRed}
\end{equation}

\noindent
Then for dominant color channel, green, and subordinate color channel, red,  we analyzed the distribution of maximum pixel intensity for each image category. 
Fig.~\ref{fig:HistMaxRed} and Fig.~\ref{fig:HistMaxGreen} display the histogram of $I_{max}^{Red}$ and $I_{max}^{Green}$ grouped by non-crystal, likely-lead, and crystal categories. 
In the subordinate red channel, almost all non-crystal images have lower maximum intensity than crystals and likely-leads. For the dominant green channel, the highest intensity value of non-crystal images spread all over the intensity scale. This indicates an overlap for the green channel between non-crystal and other categories. These figures show that the subordinate red channel could be more suitable for classifying images than the dominant green channel. 

We further analyzed the overall intensity of images. The histogram of the average gray intensity is plotted in Fig.\ref{fig:HistAvgGray}. It can be seen that the average gray intensity has similar distributions for all categories. Then we have used the threshold of Otsu's thresholding algorithm \cite{Otsu1979AHistogram} and plotted the histogram of the thresholds in Fig.\ref{fig:HistOGray}. Although Otsu's method provides better distribution, the right side of the peak for non-crystal images have overlaps with crystal and and likely-lead images.
Hence, we may state that the subordinate color channel is more promising than the dominant color channel and the overall intensity while distinguishing image categories.

\subsection{Subordinate Color Percentile Intensity Thresholding}
We have developed a percentile-based thresholding method named subordinate color percentile intensity thresholding (SPI) to segment the fluorescence microscopy images using a subordinate color channel. In general, the SPI method could generate binary images using any color channel. However, in our study, SPI creates the binary image using the intensity of the subordinate color in the protein crystallization trial images (the red color in this case). 

The SPI thresholding has three threshold values. The most important factor is the \textit{color percentile threshold}, $T_{CP}$, that satisfies the percentile intensity condition over all pixels in an image. For instance, if the percentile is chosen as $90\%$ for a channel, the percentile threshold is selected in a way that 90\% of intensity values are below this threshold. Let the color channel $C$ of an image $I$ be $I_C$. The $P^{th}$ percentile intensity threshold of the $C$ color in $I$ is computed using the histogram as: 
\begin{equation}
T_{CP} = Max(k)|\sum_{i=0}^{k} histogram(I_C,i) < P \times h \times w 
    \label{eq-T-CP}
\end{equation}

The second threshold value, $t_{min}$, is the minimum intensity requirement of a crystal image. There must be some high-intensity regions in an image of a crystal. The crystal regions should have intensity high enough compared to the background. For this purpose, SPI uses t\textsubscript{min} threshold to remove any low-intensity pixels. This threshold is based on the color-channel intensity instead of the gray-level intensity. A pixel $i$ in the image $I$ is assigned to background region if the pixel intensity for color $C$ is less than or equal to $t_{min}$ threshold: 
\begin{equation}
    I(i) = 
    \begin{cases}
    \textit{background,} & \text{if } I_{C}(i) \leq t_{min},
    \\
    \textit{evaluate further} & \text{otherwise}.
    \end{cases}
\end{equation}

The third threshold is a relaxing thresholding and denoted as t\textsubscript{max}. Sometimes, the percentile threshold could be a little high and crop parts of crystal regions. To relax the percentile threshold, t\textsubscript{max} is introduced. Even though a pixel $I(i)$ intensity in the $C$ color is less than the percentile threshold, it may be chosen as a foreground pixel if its intensity value is greater than or equal to t\textsubscript{max}.
\begin{equation}
    I(i) = 
    \begin{cases}
    \textit{foreground,} & \text{if } I_{C}(i) \geq t_{max},
    \\
    \textit{evaluate further} & \text{otherwise}.
    \end{cases}
\end{equation}

The SPI image binarization equation can be expressed as:
\begin{equation}
    I(i) = 
    \begin{cases}
    foreground, & \text{if } I_{C}(i) > t_{min} \text{ and } I_{C}(i)>T_{CP} 
    \\
    \text{} & \text{or } I_{C}(i) \geq t_{max}
    \\
    background, & \text{otherwise}.
    \end{cases}
\label{eq:SPI-condtional}
\end{equation}

The procedure of SPI thresholding is presented in Algorithm \ref{alg:SPI}. The SPI algorithm takes an RGB image $I$ as an input. The subordinate color channel $C$, percentile intensity $P$, $t_{min}$, and $t_{max}$ are the formal parameters of the algorithm. SPI output the binary image $B$ of the original image $I$. The following steps summarizes the algorithm:

\begin{itemize}
    \item Step 1 (line 1): Obtain the $I_C$ intensity matrix for the color $C$ from the original image $I$.
    \item Step 2 (line 2): Compute the $P^{th}$ percentile intensity threshold $T_{CP}$ of the color $C$ using formula (\ref{eq-T-CP}).
    \item Step 3 (line 3 -6): Create the binary image. The pixel $i$ in $I_C$ is assigned to foreground region if the color $C$ intensity $I_{C}(i)$ greater than both $t_{min}$ and $T_{CP}$ or greater than or equal to $t_{max}$. Otherwise, it is assigned to the background region (\ref{eq:SPI-condtional})).
\end{itemize}

\begin{algorithm}
\caption{SPI}
\begin{algorithmic}[1]\label{spiAlg}
 \renewcommand{\algorithmicrequire}{\textbf{Input:}}
 \renewcommand{\algorithmicensure}{\textbf{Output:}}
  \REQUIRE Original Image: \texttt{I}
  \REQUIRE Color channel: \texttt{C}
  \REQUIRE Percentile intensity: $P$
  \REQUIRE Min and max intensity thresholds of color \texttt{C}: t\texttt{\textsubscript{max}}, t\texttt{\textsubscript{min}}
  \ENSURE  Binary image: \texttt{B}
\STATE Obtain channel \texttt{I\textsubscript{C}} from \texttt{I}  under color \texttt{C} 
\STATE Compute threshold \texttt{T\textsubscript{CP}} that \texttt{P} percent of pixels in \texttt{I\textsubscript{C}} has color \texttt{C} intensity $<$ \texttt{T\textsubscript{CP}}

\STATE FOREACH pixel \texttt{i} in \texttt{I\textsubscript{C}} with intensity  \texttt{I\textsubscript{C}(i)}
\indent
\STATE ~~~IF (\texttt{I\textsubscript{C}(i)}$>$t\texttt{\textsubscript{min}} $\&$ \texttt{I\textsubscript{C}(i)}$>$\texttt{T\textsubscript{CP}}) $\|$ (\texttt{I\textsubscript{C}(i)} $>=$ t\texttt{\textsubscript{max}}) THEN
\STATE  ~~~~~~~~~Set pixel \texttt{i} as foreground
\STATE ~~~ELSE set pixel \texttt{i} as background 
\end{algorithmic}
\label{alg:SPI}
\vspace{-3pt}
\end{algorithm}

We are going to use the following notation to denote SPI thresholding of a color channel \texttt{C} at a particular percentile intensity \texttt{P}: $Sub$ followed by the color channel and the color channel followed by the percentile as SubCP. For example, $SubR90$ means SPI thresholding on the subordinate red channel (SubR) with $90^{th}$ percentile. In addition, 
$SubR90$ uses (t\textsubscript{max}) and (t\textsubscript{min}) thresholding values.

There is one significant difference between the SPI method we develop for 
subordinate color channel and the dominant green percentile thresholding method
described in  Section \ref{Background}.B. t\textsubscript{min} of dominant green percentile thresholding is based on the gray level. When we adapt  t\textsubscript{max} threshold to the original dominant green percentile intensity method, this threshold is applied for gray level. 
For example, consider a pixel \texttt{i} in dominant green channel \texttt{I\textsubscript{G}} with intensity \texttt{I\textsubscript{G}(i)} and general gray level \texttt{I\textsubscript{gray}(i)}. This pixel is set as a foreground pixel for green percentile intensity thresholding using three threshold values, (t\textsubscript{max}), (t\textsubscript{min}), and T\textsubscript{GP} if the following condition is satisfied:
$(I_{gray}(i)>t_{min} \textit{ and } I_{G}(i)>T_{GP}) \textit{ or }(I_{gray}(i)\geq t_{max})$.

\section{Experiments and Results}\label{Expt}
All experiments were performed on CR-6G fluorophore attached to proteins of interest. Our Crystal X2 fluorescence microscope is used to capture the screening samples. The image segmentation and feature extraction were implemented using MATLAB (2018a), and the classification models are built onTFL276

The experiments in this section examine the classification of protein crystallization trials based on SPI segmentation method. We analyze the performance of SPI method based on the subordinate channel and evaluate how it performs with respect to the dominant color channel. 
We evaluate its performance on classification and also provide how it can benefit from the variation of thresholding values. 
Then, we provide the overall classification performance of using the subordinate color channel for segmentation and feature extraction.
Finally, we have performed additional new wet-lab experiments and check the performance of using the subordinate color to determine its generalizability.

While evaluating the classification performance and thresholding values, we used 90\% of the image data set for training and the remaining 10\% of the data for final evaluation of the optimized model. The images were randomly split but the ratio of image categories were preserved as in stratified sampling.

\subsection{Evaluation Measures}
The objective of our research is to minimize the rate of missing a crystal, in other words, the misclassification of any crystal or leading condition  as a non-crystal. 
The evaluation measures that are used in this study are as follows:

\begin{enumerate}
\item Misclassification of both likely-leads and crystals as non-crystal rate (lower is better):
\begin{equation}
MLC = \frac{ml + mc}{l + c}\label{eq1}
\end{equation}

\item Misclassification of crystals as non-crystal rate (lower is better):
\begin{equation}
MC = \frac{ mc}{c}\label{eq2}
\end{equation}

\item Accuracy (higher is better):
\begin{equation}
ACC = \frac{ cn + cl  + cc}{n +l+c}\label{eq3}
\end{equation}
\end{enumerate}

\noindent
where 
$n$ is the number of non-crystals,
$l$ is the number of likely-leads,
$c$ is the number of crystals,
$cn$ is the number of correctly classified  non-crystals,
$cl$ is the number of correctly classified likely leads,
$cc$ is the number of correctly classified crystals,
$ml$ is the number of likely-leads  misclassified as non-crystal, and
$mc$ is the number of crystals misclassified as non-crystal.

\subsection{Performance of the Subordinate Color Channel on Classification}
In this section, we compare the performance of our SPI method using the subordinate channel to other thresholding methods relying on the dominant color channel or overall intensity. 
Hence, to experimentally investigate the effectiveness of the proposed SPI thresholding method, we trained and evaluated the performance of different classifier algorithms on the feature sets extracted from SPI and other thresholding methods.      

More specifically, we compare the SPI method using the subordinate color channel with dominant green  percentile thresholding and the Otsu's thresholding using the overall intensity. 
The Otsu's thresholding method calculates its optimum threshold value based on the assumption of bimodal histogram distribution\cite{Otsu1979AHistogram}. Otsu's thresholding does not require t\textsubscript{min} or t\textsubscript{max} parameters. For the dominant green percentile intensity thresholding method, we used four percentiles for thresholding as DomG90, DomG97, DomG99, and DomG100. The t\textsubscript{min} was set equal to 40 as reported in our previous research\cite{Dinc2014:DTbinarize,Dinc2017:Super-Thresholding,Sigdel2013:Real-Time,Sigdel2014:GeometricFeaatures,Sigdel2017:Featureanalysis} and t\textsubscript{max} was set to 255 (the highest value on grayscale). The proposed SPI method is performed on the subordinate red color channel with four percentiles for thresholding of SubR90, SubR97, SubR99, and SubR100. Initially, we set t\textsubscript{min} = 0, the lowest red intensity value, and set t\textsubscript{max} equal to 255. The TFL2756 images were segmented by each thresholding method. Then, the features were extracted from each image to create the feature set (see Section \ref{section:feature-extraction}).  
These feature sets were used to train and evaluate the classification performance. Each feature set has 52 attributes and a class label of non-crystal, likely-lead, or crystal category.  

\begin{figure*}[t!]
\centering
\minipage{1.0\textwidth}
\includegraphics[width=\linewidth]{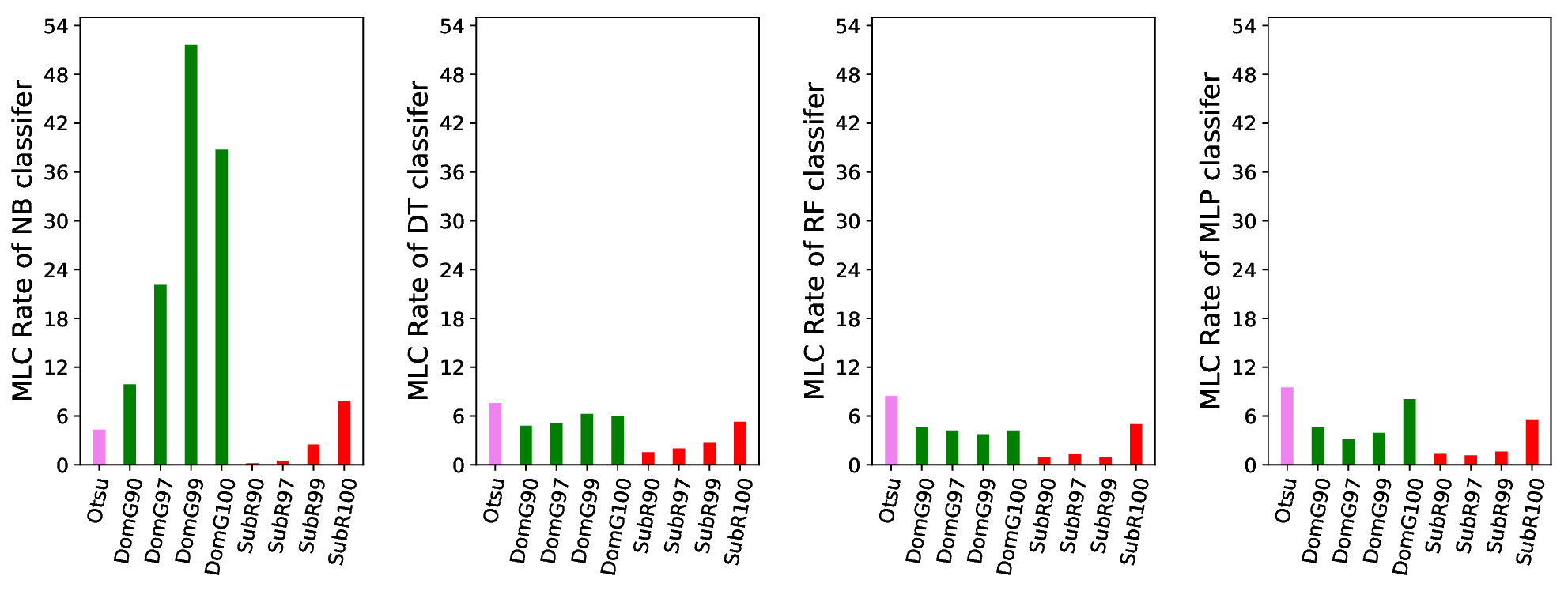}
\caption{The rate of misclassifying crystal and likely-lead as non-crystal (MLC) of NB, DT, RF, and MLP.}
\label{fig:Classiers-compare-MLC}
\endminipage\hfill
\minipage{1.0\textwidth}
\includegraphics[width=\linewidth]{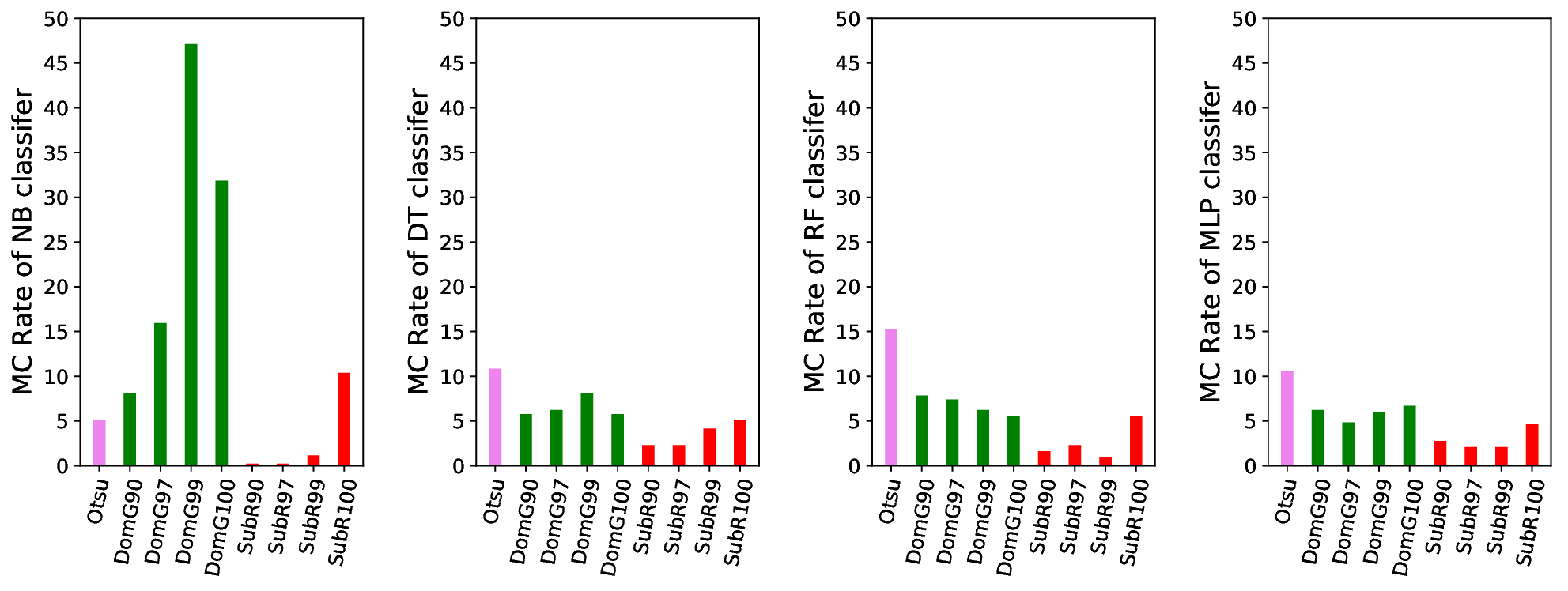}
\caption{The rate of misclassifying crystal as non-crystal (MC) of NB, DT, RF, and MLP.}
\label{fig:Classiers-compare-MC}
\endminipage\hfill
\minipage{1.0\textwidth}
\includegraphics[width=\linewidth]{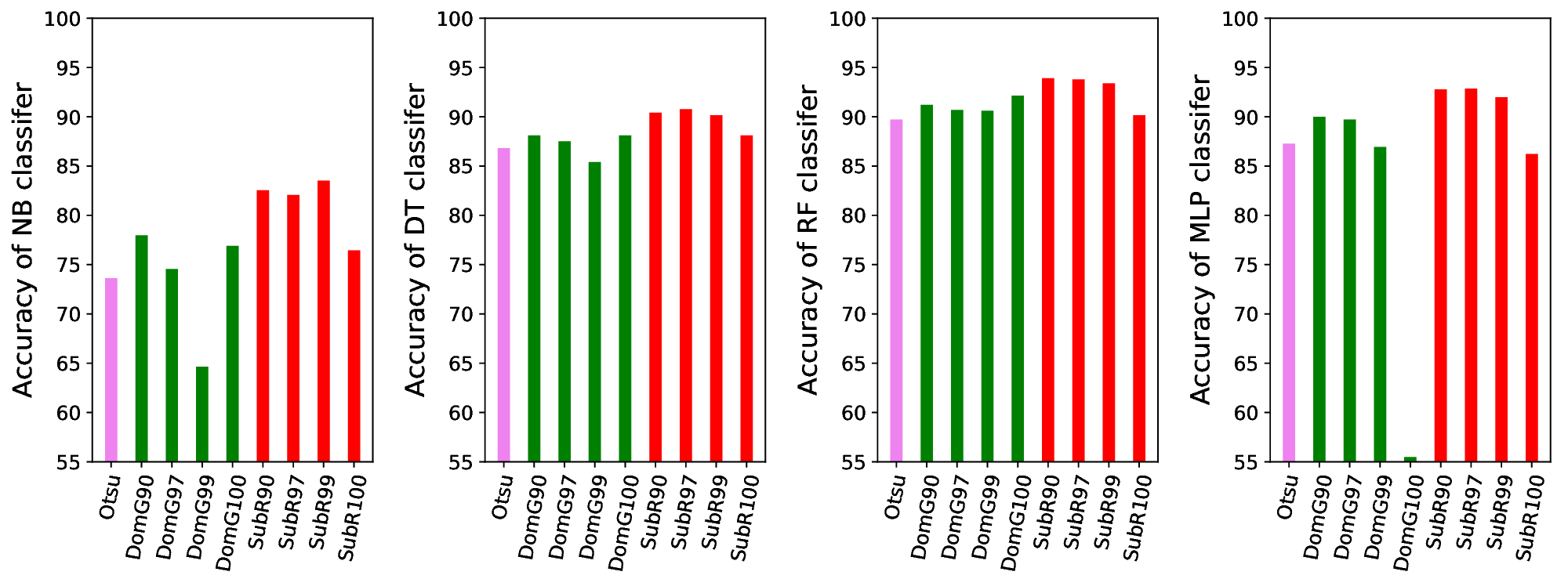}
\caption{The accuracy of NB, DT, RF, and MLP.}
\label{fig:Classiers-compare-ACC}
\endminipage\hfill
\end{figure*}

For classification models, we used four algorithms: Naive Bayes (NB), which is a classifier using Bayes theorem with assumptption that a particular feature is independent of other features; Decision Tree (DT), which we used C4.5 algorithm and the pruning condition was set as (minimum) 2 instances per leaf; Random Forest (RF), which was set to have 100 trees; and Multilayer Perceptron Neural Networks (MLP), which was constructed with one hidden layer of 27 neurons, the input layer of 52 neurons, and the output of 3 nodes.

Among the four classifiers, NB had the lowest ACC ($<$80\%) and a high MLC rate (averaging around 15.30\% for the dominant color channel using DomG90, DomG97, DomG99, and DomG100) as shown in Fig. \ref{fig:Classiers-compare-MLC} and Fig. \ref{fig:Classiers-compare-ACC}. Using the NB classifier,  SubR90 achieved MLC rate close to 0 and accuracy close to 85\%. 
Among all classifiers RF performed well for both the dominant and subordinate color channels. However, the MLC rate of SubR90 was 0.96 (below 1) whereas the MLC rate of DomG99 was only 3.75.
Using the subordinate channel not only outperformed for RF and NB, but it also performed better for MLP and DT classifiers. These results show that the superiority of using the subordinate channel is across four classifiers used in this study. 

Compared to using the dominant color channel, our segmentation method on subordinate color channel was able to the reduce the misclassification rate of likely-leads or crystals as non-crystals by the percentage of from $9.71\%$ to $2.02\%$ depending on the classifier. Similarly, the accuracy of classifiers were increased by the percentage of from $1.77\%$ to $5.53\%$. Our method reached $93.91\%$ accuracy while keeping misclassification of likely-leads and crystals as non-crystals below $1\%$. When using our method, the rate of a crystal misclassified as non-crystal (MC) is also decreased compared to the method utilizing the dominant color channel (see Fig. \ref{fig:Classiers-compare-MC}).   
Among the SPI features sets, SubR99 stands out as the best data set yielding high accuracies and low MLC rates. SubR99 results MLC rate of around 1\% with an accuracy of $\sim$93.5\% for RF and $\sim$92.0\% for MLP classifiers.

\subsection{Tunability of Thresholding Factors for the Subordinate Channel}
The goal of this experiment to determine if tuning the threshold values benefit the classification performance and shows possible improvements while binarizing the images.
To investigate the impact of threshold factors we analyze 36 pairs of $t_{min}$ and $t_{max}$ parameters on the proposed SPI algorithm. We also evaluate the dominant green percentile intensity thresholding for comparison. Based on the results of the previous experiments, we examine SubR99 and DomG99 in this section. 

There are 6 different values of intensity for each threshold factor with an interval of 32 for $t_{max}$ from 95 to 255 and 16 for $t_{min}$ from 0 to 79.    
To evaluate the threshold values, we trained two classification models, RF and MLP, on feature sets extracted for each thresholding iteration. The models were evaluated using 10-fold cross-validation. Totally, 72 models were trained and tested. 

\begin{figure}[t]
\centering
\minipage{0.5\textwidth}
\includegraphics[width=\linewidth, trim={12cm 0 12cm 1.2cm},clip]{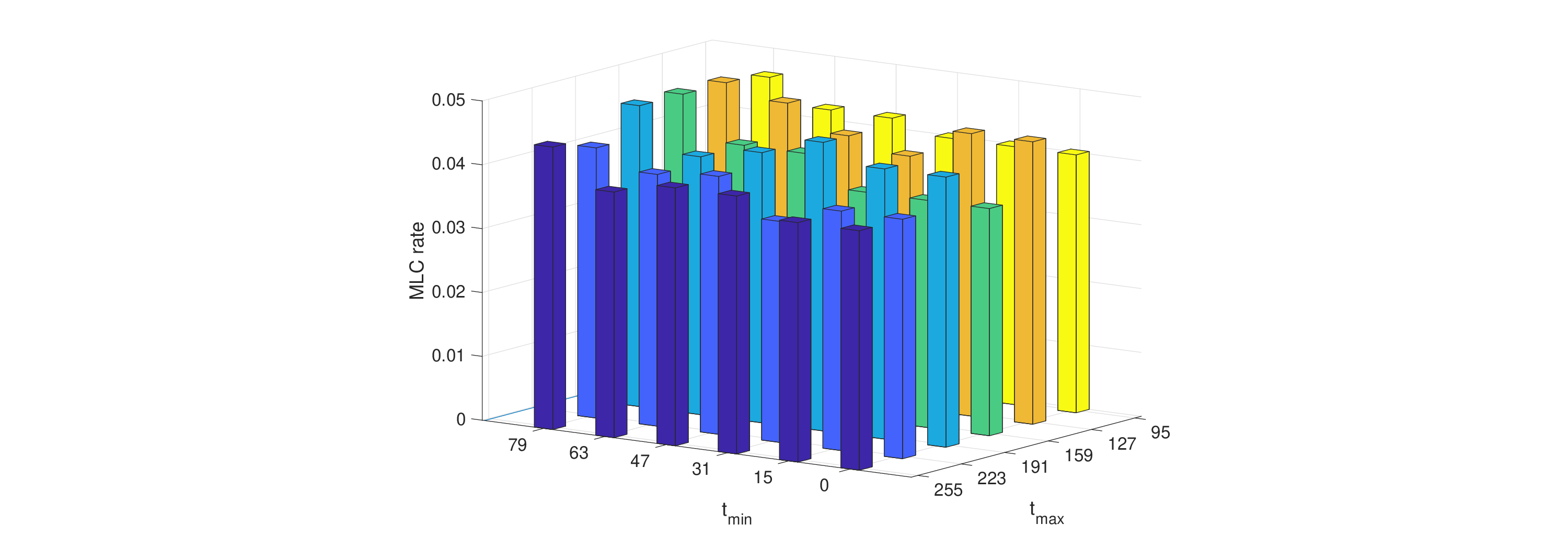}
\caption{MLC rate of RF classifier for Dom99 by varying thresholds.}
\label{fig:bar-RF-Green}
\endminipage\hfill
\minipage{0.5\textwidth}
\includegraphics[width=\linewidth, trim={12cm 0 12cm 1.2cm},clip]{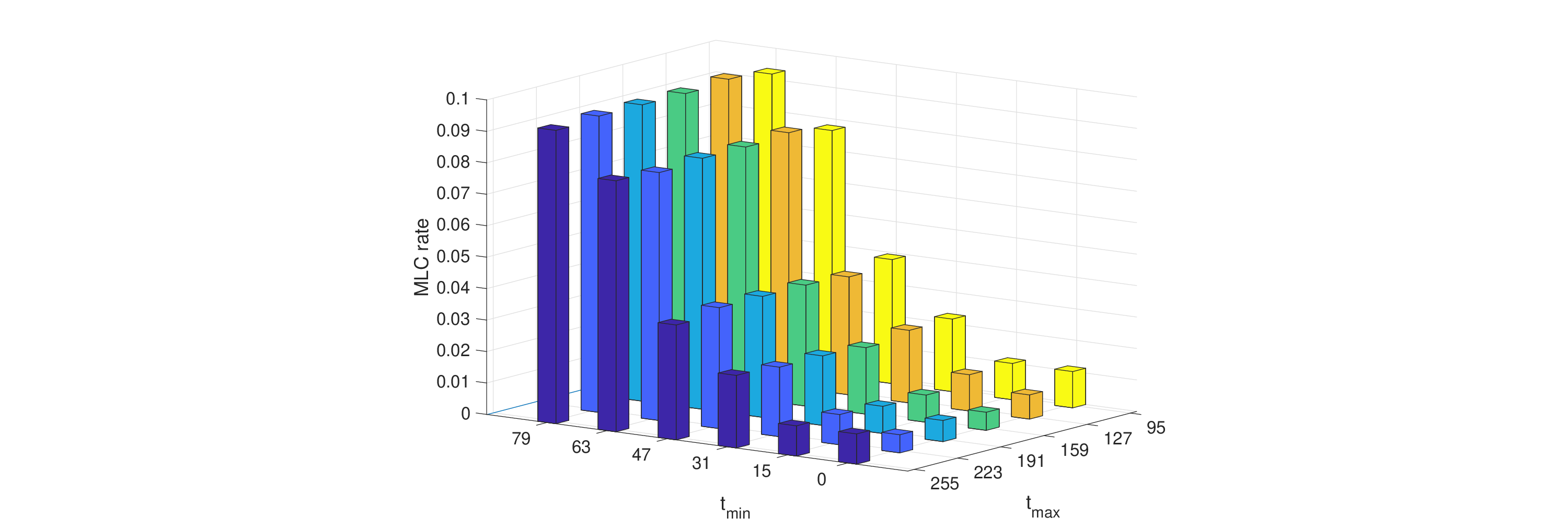}
\caption{MLC rate of RF classifier for SubR99 by varying thresholds.}
\label{fig:bar-RF-Red}
\endminipage\hfill
\vspace{-10pt}
\end{figure}

\begin{figure}[t]
\centering
\minipage{0.5\textwidth}
\includegraphics[width=\linewidth, trim={10cm 0 10cm 1.3cm},clip]{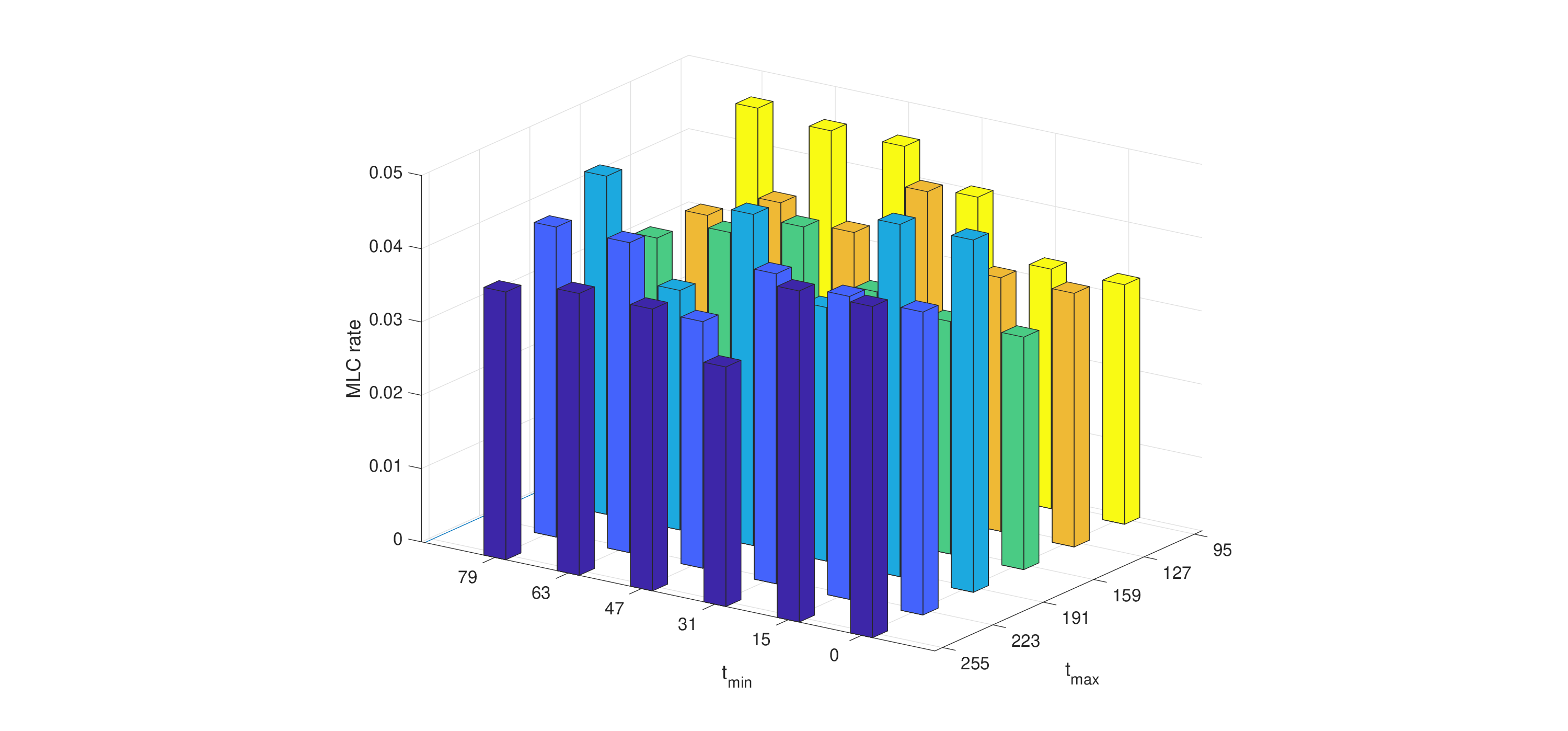}
\caption{MLC rate of MLP classifier  for DomG99 by varying thresholds.}
\label{fig:bar-MLP-Green}
\endminipage\hfill
\minipage{0.5\textwidth}
\includegraphics[width=\linewidth, trim={12cm 0 12cm 1.2cm},clip]{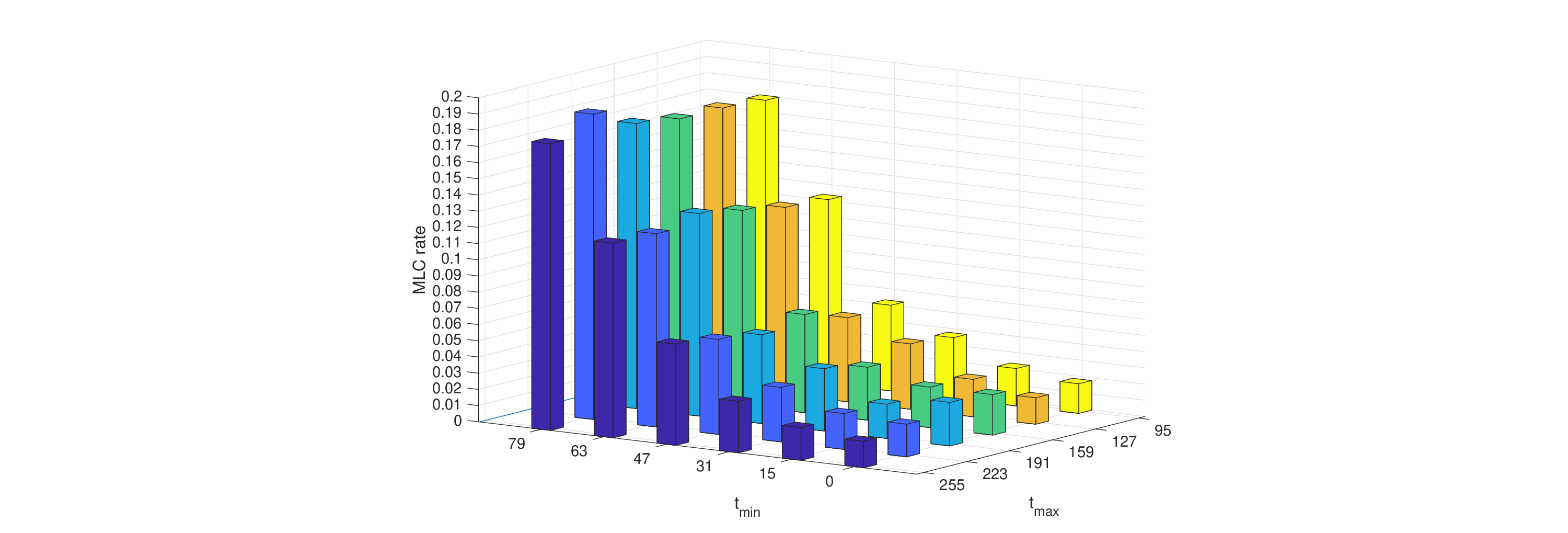}
\caption{MLC rate of MLP classifier  for SubR99 by varying thresholds.}
\label{fig:bar-MLP-Red}
\endminipage\hfill
\vspace{-15pt}
\end{figure}

Fig.~\ref{fig:bar-RF-Green} displays the MLC rate of the dominant color channel (DomG99) using the RF classifier in 3D bar chart with respect to different values of t\textsubscript{max} and t\textsubscript{min}. Fig.~\ref{fig:bar-MLP-Green} shows the similar chart for the MLP classifier. In both charts, the MLC rates of DomG99 for RF and MLP are between $0.45$ and $0.49$. It can be observed that DomG99 did not benefit much from the variation of $t_{min}$ and $t_{max}$ parameters. 

The MLC rates of the subordinate channel using RF and MLP classifiers with respect to the SPI thresholding (SubR99) are plotted in Fig.~\ref{fig:bar-RF-Red} and \ref{fig:bar-MLP-Red}. The 3D bar charts show the effects of $t_{min}$ and $t_{max}$ factors in SubR99 on reduction of  the MLC rate. The MLC rate of RF is reduced from a high rate of above 9\% to a low rate of below 1\%. The lowest MLC rate for RF classifier is just right above 0.5\% when t\textsubscript{max} = 159 and t\textsubscript{min} = 0. For the MLP classifier, the MLC rate is reduced from above 17\% to around 1\%.

These results show that the dominant color channel could not benefit from varying thresholding values, which imply that methods relying on global thresholds may not be improved much if the dominant color channel is used. On the other hand, the thresholding values had a significant positive impact on the subordinate color channel and showed the room for possible improvements while binarizing the images and extracting features.

\subsection{Overall Performance using the Subordinate Color Channel}
In this section, we experimentally tested the performance of using the subordinate color channel by picking the best classifier and best set of threshold values based on the experiment results in the previous sections.
Hence, for subordinate color channel using SPI method based on SubR99, we picked RF classifier. This classifier is trained on the feature sets extracted by SubR99 with $t_{max} = 159$ and $t_{min}$ set to 0. 

Table \ref{tabvalidation} shows the performance of SubR99 on TFL2756 data sets using 90\% hold-out, 3-fold, 5-fold, and 10-fold cross-validation. 
The average accuracy of all tests was around 94\%. When holding-out 10\% of TFL2756 data for testing and training on 90\%, the MLC and MC rates of the classification model are both 0. In this case, our model did not miss any crystal or likely-leads condition in the test images. The MLC rates were 0.57\%, 0.43\%, and 0.43\% for 3-fold, 5-fold, and 10-fold cross validation.  All the cross-validation tests have the same MC rate of 0.62\%. It can be observed that our method is stable and robust on the TFL2756 protein crystallization trial images.

\begin{table}[t]
  \centering
  \caption{Classification model testing result}
    \begin{tabular}{lcrrr}
          & \multicolumn{1}{l}{Hold-out 90\%} & \multicolumn{1}{l}{3-fold} & \multicolumn{1}{l}{5-fold} & \multicolumn{1}{l}{10-fold}\\
    ACC (\%) & 94.56 & 93.75 & 93.65 & 94.30\\
    MLC (\%) & 0     & 0.57  & 0.43  & 0.43\\
    MC (\%) & 0     & 0.62  & 0.62  & 0.62\\
\bottomrule
    \end{tabular}%
  \label{tabvalidation}%
  \\
  {\raggedright *Data table used with permission from Tran et al. \cite{Tran2017:subordinate}),  \textcopyright2017 IEEE.\par}
\end{table}%

\subsection{Evaluation of the Subordinate Channel on New Wet-lab Experiments}
\begin{figure}[t]
\centerline{\includegraphics[width=0.5\textwidth]{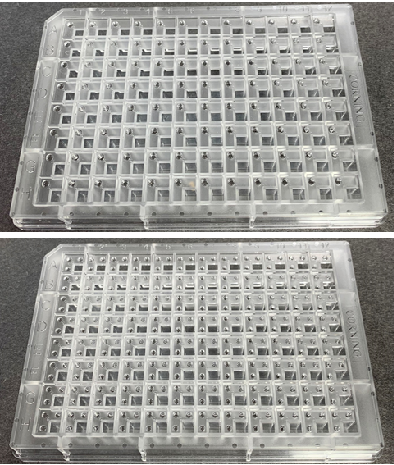}}
\caption{Sample image of the 96-well plate (top) and 288-well plate (bottom).}
\vspace{-10pt}
\label{fig:plate}
\end{figure}

To evaluate the generalizability of using the subordinate color channel, we have conducted new wet-lab experiments and collected new trial images on two proteins, 
Concanavalin A (ConA) and Ab inorganic pyrophosphatase (AbIPPase). Each protein was trace fluorescently labeled using Carboxyrhodamine 6G. Fig. \ref{fig:plate} shows the sample of the screening plates used in this experiment. 
For ConA protein, 96-well screen plate was used (8 rows by 12 columns of single sample well) giving 96 trial images. For AbIPPase, 288-well plate was used giving 288 trial images per scanning. 288-well plate has 3 sub-wells for each 96 well location (96*3=288). We used the sitting-drop vapor-diffusion for the crystallization. The trial plates were set up using standard procedures (see Section \ref{TFL procedure}).  

We used the Crystal X2 automated fluorescence microscopy to capture the images of the wells under the illumination light at 530nm. These two experiments generated 384 (96+288) trial images. The expert manually reviewed the trial images and assigned scores to each image.  
For wet-lab experiments, we have achieved an accuracy of 90.36\%, MLC of 1.42\%, and MC of 0.60\%. These results show that the classifier built using the subordinate color channel is generalizable.

\section{Discussion}\label{discussion}
Our experiment results using the subordinate color channel show that our method yields high accuracy and low misclassification rate. These results confirm our observation of the preliminary analysis that the use of subordinate color channel provides better result than the primary dominant channel. In trace fluorescent labeling, the fluorescence dyes will most likely attach to protein crystals. Since the fluorescence dye emits a wide range of light wavelengths, as shown in Fig. \ref{fig:Spectra}, the protein crystal could emit at different wavelengths other than the peak emission wavelength. 
Other chemicals such as salt crystals or precipitates do not attach to the dye molecules. However, because of free floating dye molecules or other reflective chemicals, it is possible to observe the reflection for non-crystal protein regions as well. 
Therefore, analyzing the fluorescence image beyond color channel corresponding to the illumination color may increase the robustness of the analysis.

\subsection{Comparing Binarized Images of Dominant and Subordinate Color Channels}
\begin{figure}[t]
\centerline{\includegraphics[width=0.45\textwidth]{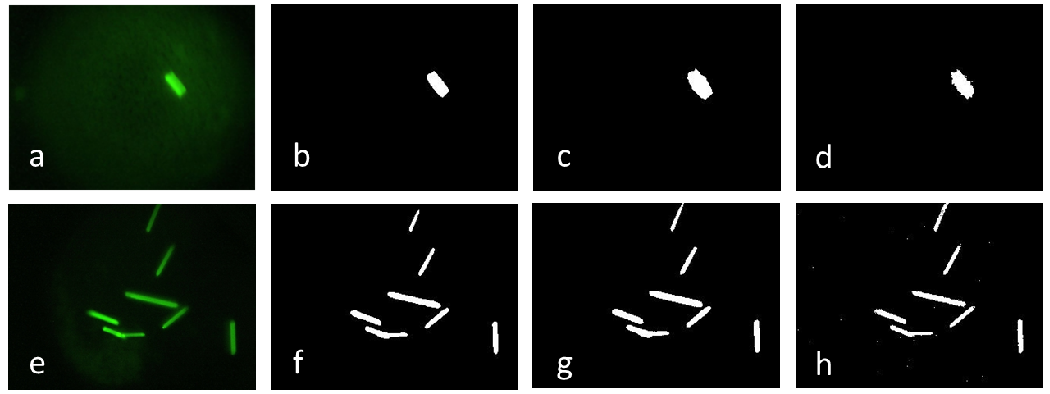}}
\caption{Binary images of crystals (a,e)  conducted by different thresholdings: Otsu (b, f); Green channel (c: DomG99, g: DomG97 with t\textsubscript{max} = 255 and t\textsubscript{min} = 40)); Red channel ( d: SubR99, h: Sub97 with t\textsubscript{max} = 255 and t\textsubscript{min} = 0).}
\vspace{-10pt}
\label{fig:SameGood}
\end{figure}

Thresholding methods can generate proper binarization if images have clearly separated regions as in Fig.~\ref{fig:SampleCategories} (h). The binarization results of Otsu, DomG99 and SubR99 thresholding methods are shown in Fig.~\ref{fig:SameGood}. Any of the thresholding methods provided good segmentation for this image.

\begin{figure}[t]
\centerline{\includegraphics[width=0.5\textwidth]{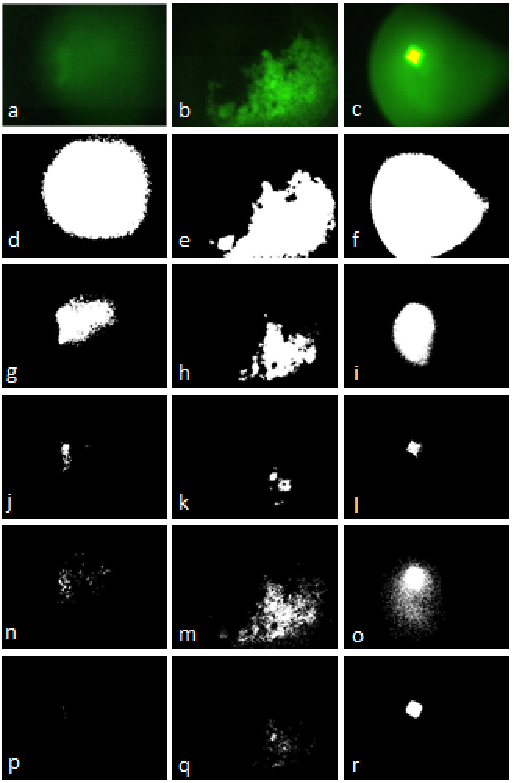}}
\caption{Binary images conducted by different thresholding: (a) original non-crystal; (b) original likely lead; (c) original crystal; (d - f) Otsu thresholding; (g-i) G90, and (j-l) G99 where t\textsubscript{max} = 255, t\textsubscript{min} = 40;
(n-o) SubR90, and (p-r) SubR99 where t\textsubscript{max} = 255, t\textsubscript{min} = 0 (Image used with permission from Tran et al. \cite{Tran2017:subordinate},  \textcopyright2017 IEEE).}
\vspace{-10pt}
\label{fig:SampleBinary}
\end{figure}

Crystal or foreground regions are not always easily separable from the background for images shown in the first row of Fig.~\ref{fig:SampleBinary}. Otsu's thresholding method generates a large foreground area that includes background or non-crystal regions as shown in Fig.~\ref{fig:SampleBinary}(d-f). While DomG90 makes a slight improvement with respect to Otsu's method by cropping some of the background or non-crystal regions as shown in Fig.~\ref{fig:SampleBinary}(g-i), its output images would be deemed incorrect. The other three methods, DomG99, SubR90, and SubR99, have produced better results. DomG99 still generates foreground regions that belong to the background (Fig.~\ref{fig:SampleBinary}(j));  however, its crystal binarization shown in Fig.~\ref{fig:SampleBinary}(l) would be acceptable. The segmentation images of SubR90  are shown in Fig.~\ref{fig:SampleBinary}(n-o). The best results are generated by SubR99 as shown in Fig.~\ref{fig:SampleBinary}(p-r) which removed non-crystal areas and background quite well as in Fig.~\ref{fig:SampleBinary}(p) and had a clear crystal region as depicted in Fig.~\ref{fig:SampleBinary}(r). It is visually observed that using the subordinate color channel generates better segmentation results.

\subsection{Misclassified Trial Images}
\begin{figure}[t]
\centerline{\includegraphics[width=0.45\textwidth]{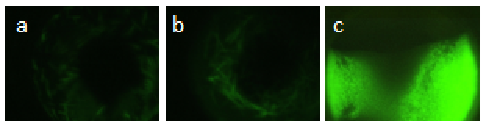}}
\caption{Sample of trial images were misclassified as non-crystal: (a-b) low intensity needle crystal; (c) birefringent precipitate likely-lead (Image used with permission from Tran et al. \cite{Tran2017:subordinate}, \textcopyright2017 IEEE).}
\vspace{-10pt}
\label{fig:MisedNeedleLikely}
\end{figure}

Even though the experiment results show that our method has high accuracy and low misclassification rate, there are still leading crystallization conditions that are misclassified as non-crystals. Fig.~\ref{fig:MisedNeedleLikely} displays some images of likely-lead and crystal that were misclassified as non-crystal images. These misclassification outcomes due to low intensity images or large bright background regions. Using proper lighting and in-focus images could improve the image quality, and methods such as FocusALL \cite{Sigdel2016:FocusAll} could be used to generate in-focus images from a stack of images. 

\subsection{Building Multiple Models for Varying Intensities.} 
Our model is built based on a training set that includes images that have varying intensities. Instead of building one model (i.e., one threshold parameter set), several models could be built for images whose average intensities are within a range. Whenever an input trial image is provided, the learning model for the corresponding intensity can be used to improve performance. Normally, this would not be needed; however, there could be lighting problems during image acquisition and captured images may have different intensities than usual. 

\subsection{Evaluation of Thresholds} There are three thresholds used in our method: percentile threshold, $t_{min}$ and $t_{max}$. 
Fig. \ref{fig:bar-RF-Red} and Fig. \ref{fig:bar-MLP-Red} show that MLC rate reduces as $t_{min}$ reduces. For the same $t_{min}$ value, different $t_{max}$ values yield the same MLC rate. 
It may be observed that we may choose $t_{min}$ close to 0, and pick a $t_{max}$ value. Our method may only use $t_{min}$ and the percentile threshold, $T_{CP}$ and can get the same performance.

\vspace{-10pt}
\section{Conclusion}\label{Conclusion}

In this paper, we introduced the use of a subordinate channel besides the color channel corresponding to the dominant wavelength in the emission spectra of fluorescence dye for classifying protein crystallization trial images.
We have developed a segmentation method based on the intensity of the subordinate color channel using three thresholds.
By using the subordinate channel, we were able to reduce the misclassification of crystals below 1\%. 
Our results indicate that using the subordinate channel in the emission spectra could be more helpful in distinguishing crystals than the color channel corresponding to the dominant wavelength. Using the subordinate channel had higher accuracy and lower MLC rates when four classifier models were used. Moreover, the segmentation based on the subordinate color channel also benefited from the variation of threshold values.

\vspace{-10pt}
\section*{Acknowledgment}
This research was supported by National
Institutes of Health (GM116283) grant.

This paper is an extension of our previous work "T. X. Tran and R. S. Aygun and M. L. Pusey, Classifying protein crystallization trial images using subordinate color channel, in 2017 IEEE Int. Conf. on Bioinformatics and Biomedicine (BIBM), Nov. 2017, page 1546-1553, DOI 10.1109/BIBM.2017.8217890". \textcopyright2017 IEEE. Reprinted, with permission from T. X. Tran and R. S. Aygun and M. L. Pusey, Classifying protein crystallization trial images using subordinate color channel, in 2017 IEEE Int. Conf. on Bioinformatics and Biomedicine (BIBM), Nov. 2017, page 1546-1553, DOI 10.1109/BIBM.2017.8217890.  
\vspace{-10pt}

\bibliographystyle{spbasic} 
\bibliography{BIBTeX.bib}

\section*{Author Biography}
\begin{wrapfigure}{l}{0.2\columnwidth}
\begin{center}
   \includegraphics[width=0.2\columnwidth]{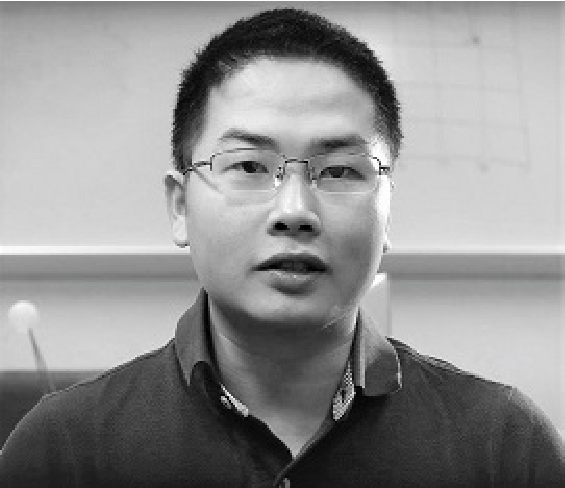}
\end{center}
\end{wrapfigure}

\textbf{Truong X. Tran} received his Bachelor's degree major in Electronic and Telecommunication Engineering, minor in Biomedical Electronic Engineering, from Hanoi University of Science and Technology, Vietnam in 2010, the M.S. degree in Computer Science from Arkansas State University, AR, USA in 2012. He is pursuing Doctoral degree in Computer Science at the University of Alabama in Huntsville, AL, USA. He works in Data analysis for protein crystallization research project. His research interests include machine learning and data mining.
\newline

\begin{wrapfigure}{l}{0.2\columnwidth}
\begin{center}
    \includegraphics[width=0.2\columnwidth]{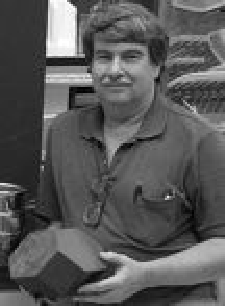}
\end{center}
\end{wrapfigure}

\textbf{Marc L. Pusey} is a research scientist working at iXpressGenes, Inc., Huntsville Alabama.  Dr. Pusey received his Ph.D. in Biochemistry from the University of Miami, Fl., then did post-doctoral research at the University of Minnesota.  He moved to Huntsville in 1985, and worked at NASA/MSFC for the next 23 years in the new field of protein crystal growth.  After retirement from NASA, he obtained his current position at iXpressGenes with the focus of his research being improved methods for protein crystal screening, crystal detection, and crystallization condition identification using visible fluorescence methods.
\newline

\begin{wrapfigure}{l}{0.2\columnwidth}
\begin{center}
    \includegraphics[width=0.2\columnwidth]{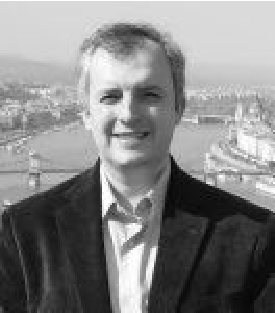}
\end{center}
\end{wrapfigure}

\textbf{Ramazan S Ayg\"{u}n:} received the B.S. degree in computer engineering from Bilkent University, Ankara, Turkey in 1996, the M.S. degree from Middle East Technical University, Ankara in 1998, and the Ph.D. degree in computer science and engineering from State University of New York at Buffalo in 2003. He is currently an Associate Professor in Computer Science Department, University of Alabama in Huntsville. His research interests include protein crystallization image analysis, data mining, image and video processing, spatio-temporal indexing and querying, multimedia databases, semantic computing, multimedia networking, and multimedia synchronization.

\end{document}